\documentclass[letterpaper, 10pt, conference]{ieeeconf}

\let\labelindent\relax

\usepackage{bm}
\usepackage{svg}
\usepackage{soul}
\usepackage{cite}
\usepackage{tikz}
\usepackage{color}
\usepackage{times}
\usepackage{braket}
\usepackage{diagbox}
\usepackage{multirow}
\usepackage{multicol}
\usepackage{listings}
\usepackage{subfiles}
\usepackage{graphicx}
\usepackage{titlesec}
\usepackage{algorithm}
\usepackage{subcaption}
\usepackage[utf8]{inputenc}
\usepackage[inline]{enumitem}
\usepackage{xcolor, colortbl}
\usepackage{booktabs,ragged2e}
\usepackage[hidelinks]{hyperref}
\usepackage[noend]{algpseudocode}
\usepackage[flushleft]{threeparttable}
\usepackage{amsmath, amssymb, amsfonts}
\usepackage[font=small, aboveskip=0.5\baselineskip, belowskip=-1\baselineskip]{caption}

\makeatletter
\newcommand\footnoteref[1]{\protected@xdef\@thefnmark{\ref{#1}}\@footnotemark}
\makeatother

\hypersetup{
  colorlinks=true,
  linkcolor=black,     
  urlcolor=blue
}

\graphicspath{{./Figures/}}

\usetikzlibrary{arrows, patterns}

\definecolor{blue}{rgb}{0.25, 0.41, 0.88}
\definecolor{red}{rgb}{1.0, 0.11, 0.0}

\newcommand{\id}[1]{\mbox{\it #1}}
\newcommand{\mypara}[1]{\medskip\noindent\textbf{#1}:\hspace*{1ex}}
\newcommand{\relframe}[5]{\sideset{_{#2}^{#3}}{_{#5}^{#4}}{\mathop#1}}

\titlespacing\section{0pt}{4pt}{4pt}
\titlespacing\subsection{0pt}{2pt}{2pt}
\titlespacing\subsubsection{0pt}{2pt}{2pt}

\newcommand{\ignore}[1]{}

\IEEEoverridecommandlockouts

\title{\LARGE \bf
Transferring Kinesthetic Demonstrations across Diverse Objects for Manipulation Planning\thanks{This work was supported in part by the US Department of Defense through ALSRP under award No. HT94252410098 and a SBU LINCATS award.}
}

\author{Dibyendu Das$^{1}$, Aditya Patankar$^{2}$, Nilanjan Chakraborty$^{2}$, C. R. Ramakrishnan$^{1}$, and I. V. Ramakrishnan$^{1}$
\thanks{$^{1}$Dept. of Computer Science, 
        Stony Brook University, USA.
        {\tt\small \{didas, cram, ram\}@cs.stonybrook.edu}}%
\thanks{$^{2}$Dept. of Mech. Engg., 
        Stony Brook University, USA.
        {\tt\small \{aditya.patankar, nilanjan.chakraborty\} @stonybrook.edu}}%
}

\begin{document}

\maketitle

\pagestyle{empty} 

\begin{abstract}
Given a demonstration of a complex manipulation task, such as pouring liquid from one container to another, we seek to generate a motion plan for a new task instance involving objects with different geometries. This is nontrivial since we need to simultaneously ensure that the implicit motion constraints are satisfied (glass held upright while moving), that the motion is collision-free, and that the task is successful (e.g., liquid is poured into the target container). We solve this problem by identifying the positions of critical locations and associating a reference frame (called \emph{motion transfer frames}) on the manipulated object and the target, selected based on their geometries and the task at hand. By tracking and transferring the path of the motion transfer frames, we generate motion plans for arbitrary task instances with objects of different geometries and poses. We show results from simulation as well as robot experiments on physical objects to evaluate the effectiveness of our solution. A video supplement is available on YouTube: \href{https://youtu.be/RuG9zMXnfR8}{https://youtu.be/RuG9zMXnfR8}

\ignore{
In many complex manipulation tasks, the same task can be performed with different objects. For example, pouring is a complex manipulation task, with constrained end-effector motion, which can be done with containers of various shapes. Intuitively, we understand that the pouring task requires a similar motion for a range of different geometries of the source and/or destination containers. However, the problem of transferring a demonstration for a given task across different objects, all of which can be used to do the same task, is an open problem. In this paper, we study the problem of using one kinesthetic demonstration of a complex manipulation task on a given set of objects to generate manipulation plans for other sets of objects, assuming that we know that the task can be performed with the objects. We use a coordinate-invariant representation of a complex manipulation task as a sequence of constant screws, which we extract from the kinesthetic demonstration. Using this representation of a complex manipulation task as a sequence of constant screws and the geometry of the objects, we present a method to generate plans for new task instances with different objects. We present experimental results showing the effectiveness of our approach using pouring as an example task.}
\end{abstract}
\section{Introduction}
\label{sec:intro}

\ignore{
Outline:

Manipulation plans from kinesthetic demonstrations: complete tasks by mimicking the motion in a demonstration.

Which motion to mimic?
- Following the end effector pose does not generalize to different object geometries.
- Follow the "leading edge" or "point of contact": C-frames
- The assignment of c-frames may depend on the kind of task.
- Heuristics for c-frame assignment for pouring and scooping tasks.  
-- Theorems characterizing the effectiveness of the heuristics.
}

\mypara{The Context}
\textit{Complex manipulation tasks} such as pouring, scooping, etc. have implicit constraints on the motion of a robot's end-effector which are hard to formally describe a priori. 
Learning from Demonstration (LfD)~\cite{chernova2014robot,billard2016learning,calinon2007onlearning,calinon2010learning} is a popular approach, where manipulation skills are learned from demonstrations provided by human teachers. Demonstrations may be provided by using specialized hardware interfaces~\cite{chi2024universal,duan2023ar2}, via teleoperation~\cite{wu2023gello,fu2024mobile}, via videos~\cite{yang2019learning,wu2020squirl}, or \emph{Kinesthetically} by manually moving the robot arm. The constraints that characterize complex manipulation tasks are implicit in the demonstrations. Motion plans for an arbitrary task instance (characterized by poses of objects relevant to that task) have been shown to be generated based on a single kinesthetic demonstration~\cite{mahalingam2023human}. In that work, the motion of the robot's end-effector in a demonstration is represented by a sequence of constant screws in $SE(3)$. The constant screws are then transferred to a new task instance and a joint space path is computed using Screw Linear Interpolation (ScLERP) and Jacobian pseudo-inverse~\cite{sarker2020screw}.  At a high level, this ensures that the end-effector motion for the task instance ``follows'' the motion in the demonstration.

\begin{figure}[!t]
    \centering
    \includegraphics[width=\linewidth]{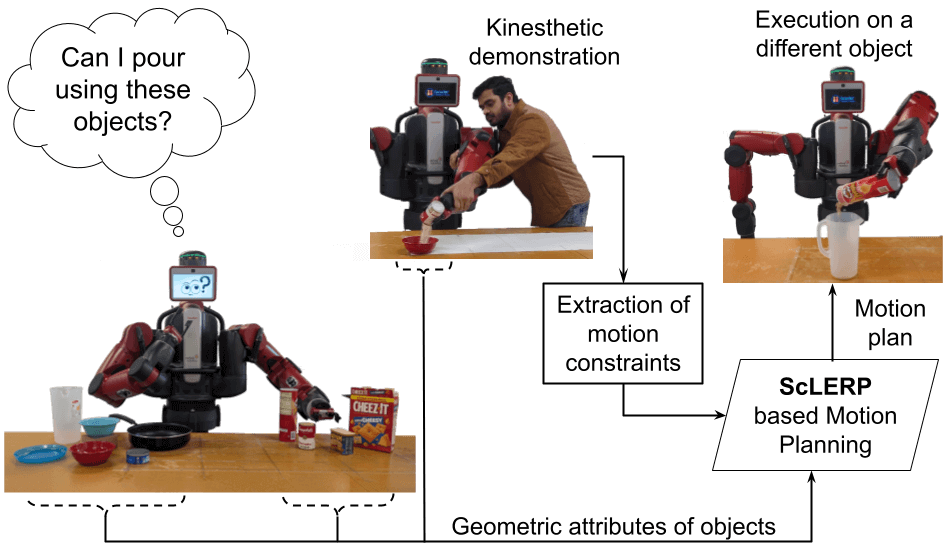}
    \caption{Example scenario and problem setting: demonstration of pouring from $\id{soup\_can}$ to a $\id{bowl}$ (middle). Motion plan generated for a new task instance, pouring from a $\id{Pringles\_can}$ into a $\id{pitcher}$ using the technique described in the paper.}
    \label{fig:pipeline}
    \vspace{\baselineskip}
\end{figure}

\mypara{The Problem} 
There are three key aspects that need to be taken into account to successfully execute complex manipulation tasks. Consider the task of pouring liquid from a glass into a bowl, as shown in Fig.~\ref{fig:pipeline}.  Firstly, the end-effector motion constraints corresponding to the task should be satisfied; for example, the glass should be held upright while moving. Secondly, the contents of the glass should be poured into the bowl, and last but not least, the glass should not collide with the bowl or other objects in the environment during motion. Although the objects did not collide during the demonstration and the liquid was poured correctly inside, this may not necessarily be the case for manipulation plans for geometrically different objects generated using the same demonstration.
\ignore{Thus when using a kinesthetic demonstration, provided using one set of objects, to generate a manipulation plan for the same task with geometrically different objects, we need to take into consideration the three key aspects described above.}
More precisely, the problem addressed in this paper is:
\emph{Given a complex manipulation task, a kinesthetic demonstration for the task on one set of objects of known geometry, compute a manipulation plan for the new task instance with a possibly different  set of objects with known geometry, and with different initial, and final poses.}

\mypara{Solution Approach and Contributions}
The key insight is that if we transfer the screws representing the motion constraints using the end-effector pose (as done in~\cite{mahalingam2023human}), we do not consider the geometry of the objects involved in the manipulation, and thus the resulting motion plan may result in collisions, or the task execution may not achieve the intended result (e.g., in pouring, one may pour outside the receiving container). We therefore identify easily computable \emph{critical locations} on the task-relevant objects and associate a reference frame (which we call \emph{motion transfer frame}) with them. The location and orientation of the motion transfer frames depend on the geometry of the objects and the task. Using the screw-geometric structure of motion to extract the motion constraints, representing them in the motion transfer frames of the demonstration objects and transferring them to the motion transfer frames of the new objects in a new task instance, we generate manipulation plans for objects different from those used in the demonstration. \emph{This novel approach for generating manipulation plans for complex manipulation tasks across different objects using a single kinesthetic demonstration is the primary contribution of this paper}. We present results for pouring, generating motion plans from a single demonstration for task instances on new objects with different geometries. We report simulation results which characterize the effectiveness of our techniques and robot experiments which validate our approach.

\section{Related Work}

\subsection{Manipulation Plan Generation using Demonstrations}

This work presents a kinesthetic demonstration-based approach for manipulation task planning, based on the idea that a single demonstration can be applied to objects of varying shapes and sizes if their geometric information is known. Although previous methods~\cite{de2010invariant, vochten2019generalizing} use coordinate-free shape descriptors of rigid body motion that compactly describe a given demonstration on an object and transfer them to new instances with new initial and goal poses of the object, they do not explicitly extract motion constraints of the task. As a result, these methods may not be generalized to complex tasks such as scooping and pouring, where constraints evolve throughout the motion.

Approaches such as Dynamical Movement Primitives (DMP)~\cite{ijspeert2013dynamical, hersch2008dynamical, pastor2009learning, saveriano2019merging} have been proposed for the generation of motion from a single demonstration. While these methods are elegant and bio-inspired, they do not explicitly account for task constraints and are not coordinate-invariant, making their performance dependent on the coordinate system of the end-effector. As a result, they struggle to generalize to task space regions beyond the demonstration~\cite{vochten2019generalizing}. In contrast, our approach explicitly extracts task constraints as constant screw motions, which are inherently coordinate-invariant.

There are also probabilistic approaches for Learning from Demonstration~\cite{billard2016learning}, e.g., those using Gaussian Mixture Models (GMM)~\cite{calinon2007learning} or Hidden Markov Models (HMM)~\cite{calinon2010learning,calinon2011encoding} but they require multiple demonstrations, and they do not extract the motion constraints of complex tasks. 

In contrast, our approach aims to extract the motion constraints using a single demonstration and apply them to different objects.

This work builds on our prior works~\cite{mahalingam2023human,laha2021point,laha2022coordinate,sarker2020screw}. The key distinction is that we extract the task constraints from the kinesthetic demonstration and transfer them to the new task instance on geometrically different objects. This allows us to reuse the same demonstration across many different objects of varying shapes and sizes.

\subsection{Manipulation Plan Generation across Geometrically Different Objects}
The problem of learning manipulation skills from demonstrations and transferring them to differently shaped objects has been studied in~\cite{thompson2021shape, gao2021kpam, wen2022you, manuelli2022kpam, gao2023k, gao2024bikvil}. As mentioned in these works, certain aspects of an object's geometry dictate the successful execution of a particular task, and the key to effective generalization is to identify these aspects. These functional aspects are captured either as key points on the object~\cite{gao2021kpam, wen2022you, manuelli2022kpam, gao2023k, gao2024bikvil} or as a learned embedding of a shape~\cite{thompson2021shape}.

Some of the existing works that employ key point-based techniques are either for pick-and-place tasks~\cite{gao2021kpam, wen2022you, manuelli2022kpam} or for very small amounts of motion, as is typical in assembly tasks (less than an inch)~\cite{wen2022you}. Manipulation plan generation across geometrically distinct objects for tasks like pouring~\cite{thompson2021shape, gao2023k, gao2024bikvil} and scooping~\cite{thompson2021shape} has been studied previously. The authors in~\cite{thompson2021shape} define the shape of an object in terms of a 3D point cloud obtained in simulation. They estimate the variation in shape within a category, which is then used to create shape embedding. The authors claim that this embedding is robust to intra-category variations. However, multiple demonstrations  are required for effective generalization within a particular category.

While we also utilize \emph{critical locations} of an object's geometry for the task of pouring, our approach differs from~\cite{gao2023k}, which requires multiple \emph{video} demonstrations for effective generalization to geometrically diverse objects. Moreover, the authors in~\cite{gao2023k} learn the motion of the key points using Via-Point Movement Primitives~\cite{zhou2019learning}, inspired by DMPs. Similar to DMPs, VMPs do not explicitly account for task constraints and are not coordinate invariant.

In contrast, we use a screw-geometric representation of task constraints, which is coordinate invariant, and by transferring them to the \emph{critical locations} of the object's geometry, we show that we can achieve a successful generalization to geometrically different objects. 



\section{Problem Formulation}
\label{problem-formulation}

In this section, we formalize the problem of using one demonstration of a manipulation task to perform the same task on objects other than those used in the demonstration itself. Before formally describing the problem, we will define some of the required concepts used in the rest of the paper.

\subsection{Task-Relevant Objects and Task Instance}

\begin{figure}[!h]
    \centering
    \includegraphics[scale=0.14]{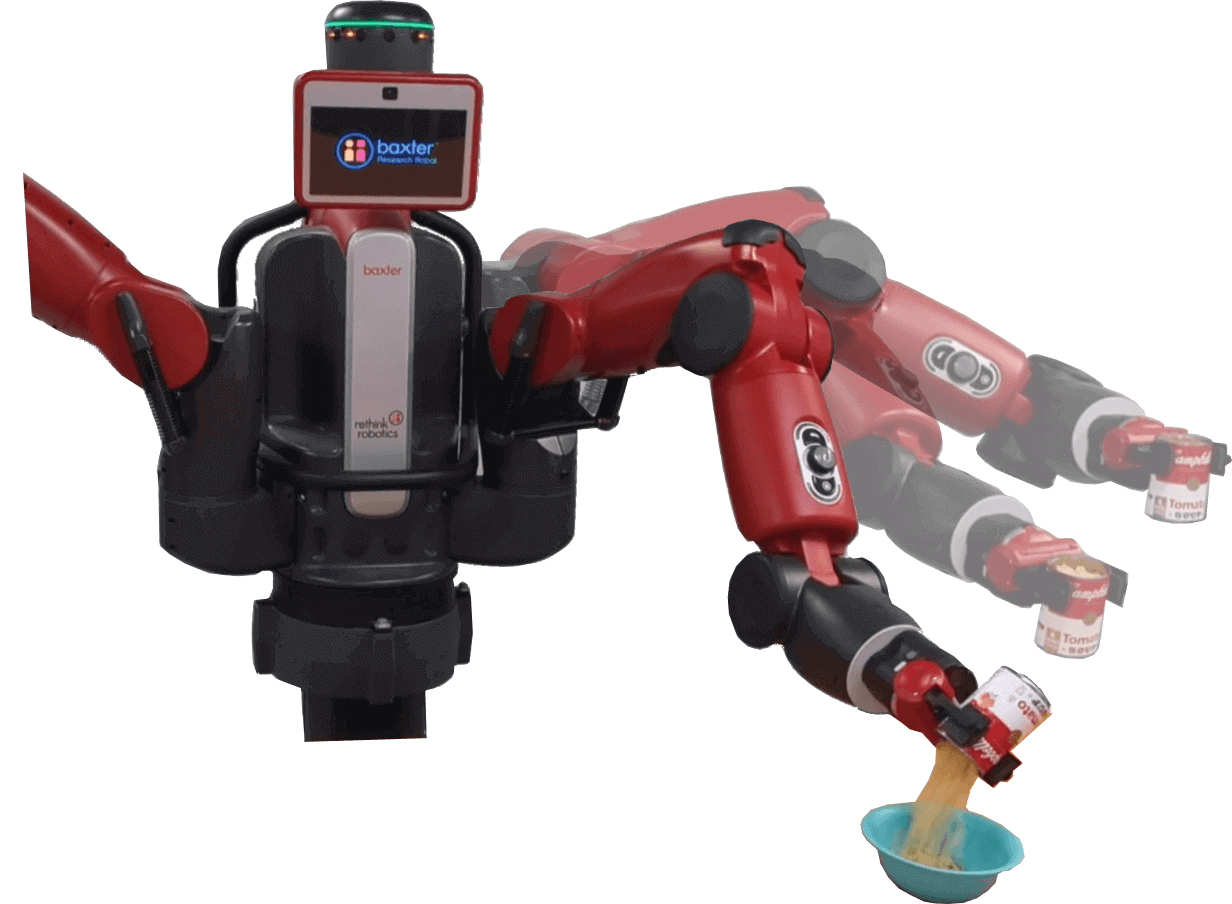}
    \caption{Pouring task involves pouring the content (secondary object) from the soup can (primary object) into the bowl (passive object).}
    \label{fig:pouring_example}
    \vspace{0.5\baselineskip}
\end{figure}

Objects whose configuration affects the manipulation plans for a task or whose poses are changed by the task, are called \emph{task-relevant objects}. Consider the task of pouring sugar from one container into another as shown in Fig.~\ref{fig:pouring_example}. The objects relevant to the task are the grains of sugar being transferred, the source and destination containers. We divide task-relevant objects into the following three categories:

\begin{description}
  \item[Primary] objects are directly manipulated by the robot's gripper\footnote{Without loss of generality, and for clarity of presentation, we  consider only those tasks that are performed with a single robot arm. We also assume that the robot gripper can hold one object at a time.}. Each task (e.g., \id{pour}, \id{scoop}) is associated with a set of primary objects, e.g., \{\id{spoon}, \id{soup\_can}\}. Each task instance directly manipulates one object from the set of primary objects (e.g. ``the \id{soup\_can} in hand").

  \item[Secondary] objects are not directly manipulated by the robot but their configuration changes during the task execution. An example is sugar grains in the pouring task; explicitly modeling their configuration is not required for manipulation planning.

  \item[Passive] objects are not directly manipulated by the robot but their poses affect the manipulation plan for task execution. An example is the \id{bowl} in the pouring task, whose poses are important for determining the start and end configurations of the robot's end-effector while performing the task.
\end{description}

\mypara{Notation}
We will use $\mathbf{g}$ to represent the pose. The notation $\relframe{\mathbf{g}}{}{}{}{AB}$ will represent the pose of the reference frame \mbox{$\{B\}$} with respect to \mbox{$\{A\}$}. When \mbox{$\{A\}$} is the world frame \mbox{$\{W\}$}, we will omit it to reduce notational clutter, i.e., we will write $\relframe{\mathbf{g}}{}{}{}{WB}$ as $\relframe{\mathbf{g}}{}{}{}{B}$. The subscripts $r$ and $s$ will be used for variables or reference frames corresponding to the \emph{primary} and \emph{passive} objects, respectively. The superscript $d$ placed on the top left will be for variables or reference frames corresponding to a demonstration, and the superscript $n$ placed on the top left will correspond to a new task. In some cases, $d$ and $n$ are used as subscripts to denote variables of interest for demonstration and new tasks.


\mypara{Task Instance}
Let $\relframe{\mathbf{g}}{}{}{}{B_r}, \relframe{\mathbf{g}}{}{}{}{B_s} \in SE(3)$ be the poses of the \emph{base reference frames} (typically located at the bottom-center of the object) associated with the primary and the passive objects, respectively. Let
$\relframe{\mathcal{O}}{}{}{}{r}, \relframe{\mathcal{O}}{}{}{}{s}$ be the \emph{geometric attributes} of the primary and the passive objects, respectively. The geometric attributes of interest depend on the task. For example, for pouring, the geometric attributes of interest are the height of the objects and the boundary curve that forms the lip (or rim) of the objects.
We define a \emph{task instance} for any given task (like \id{pour}, \id{scoop}, etc.) using the pose and geometric information of the primary and passive objects, as a tuple $t = \left(\relframe{\mathbf{g}}{}{}{}{B_r}, \relframe{\mathcal{O}}{}{}{}{r}, \relframe{\mathbf{g}}{}{}{}{B_s}, \relframe{\mathcal{O}}{}{}{}{s}\right)$.

\begin{figure*}[!ht]
    \centering
    \begin{subfigure}[t]{0.5\textwidth}
        \centering
        \includegraphics[width=\linewidth]{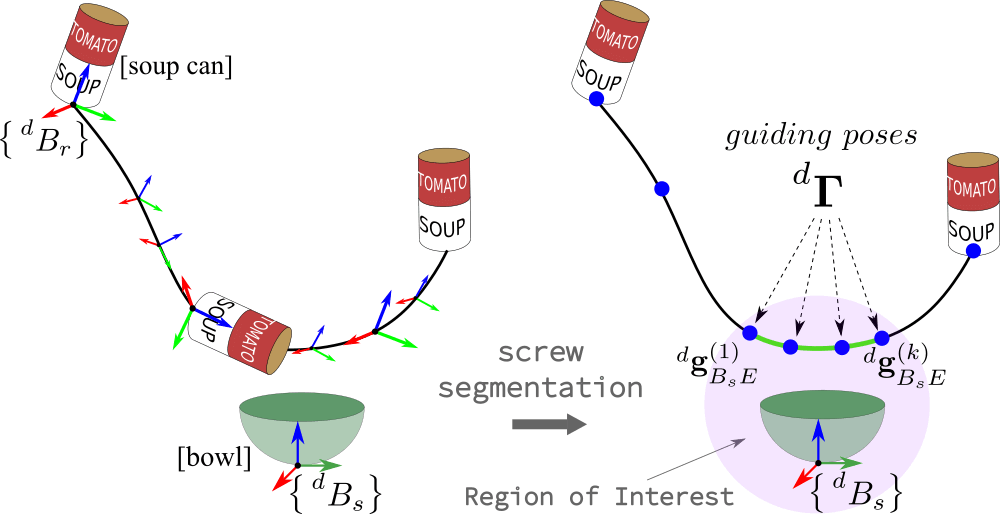}
        \caption{Motion segmentation}
        \label{fig:motion_segmentation}
    \end{subfigure}%
    ~
    \begin{subfigure}[t]{0.48\textwidth}
        \centering
        \begin{subfigure}[t]{0.5\linewidth}
            \centering
            \includegraphics[width=\linewidth]{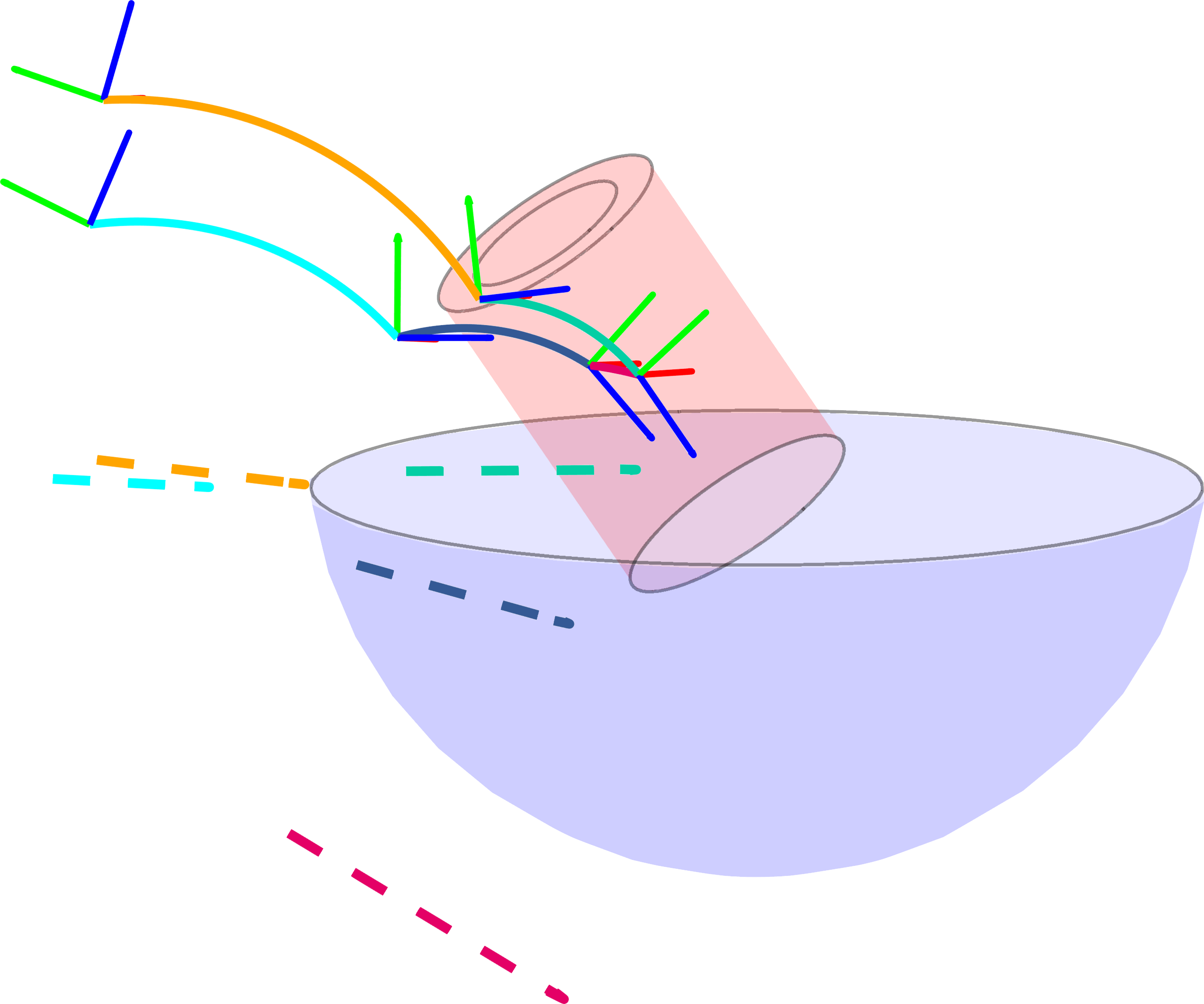}
            \caption*{Pouring from a $\id{soup\_can}$}
        \end{subfigure}%
        ~
        \begin{subfigure}[t]{0.5\linewidth}
            \centering
            \includegraphics[width=\linewidth]{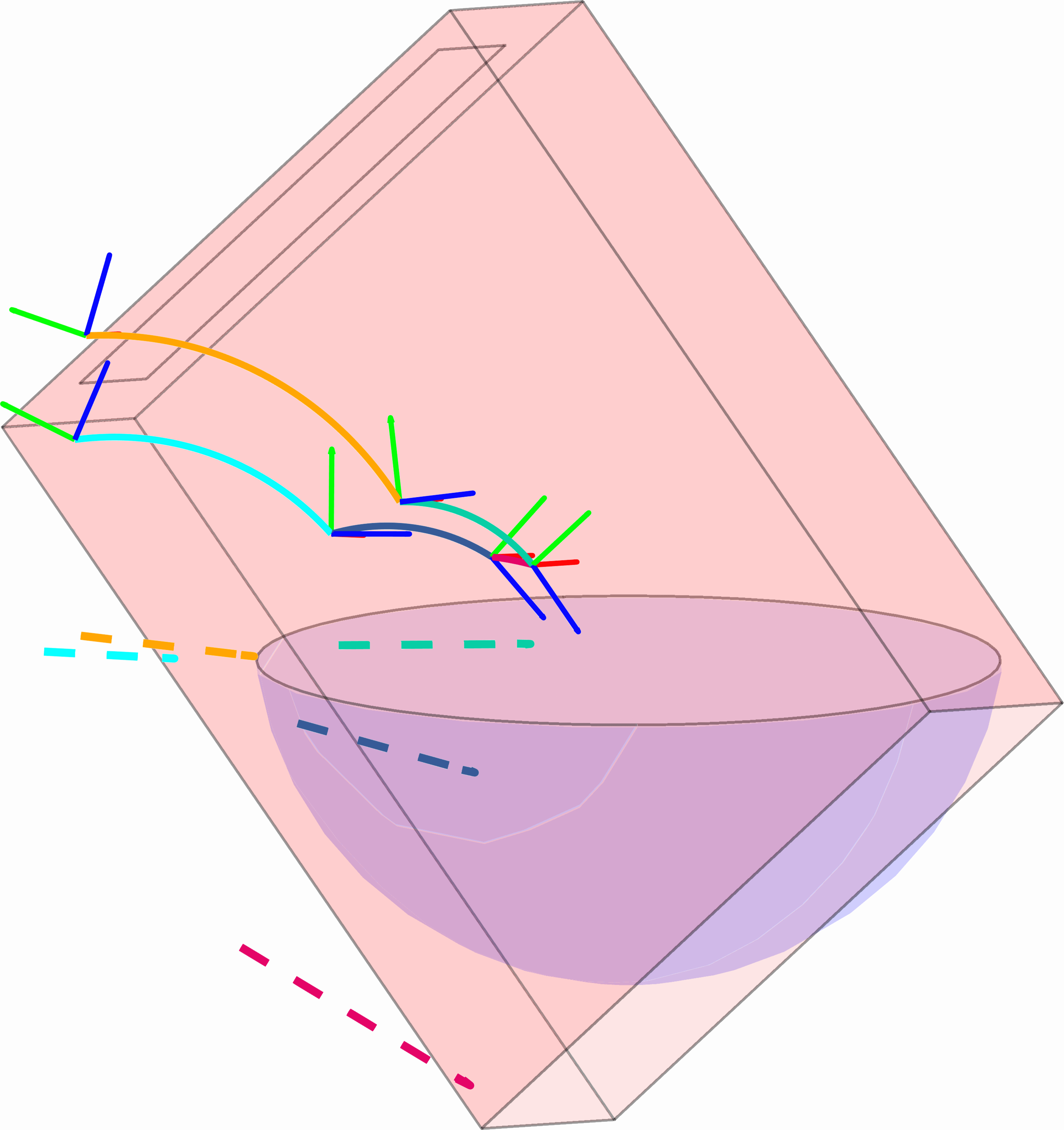}
            \caption*{Pouring from a $\id{box}$}
        \end{subfigure}
        \newline
        \caption{Screw extraction}
        \label{fig:screw_extraction}
    \end{subfigure}
    \newline\newline
    \caption{(\textbf{a}) Schematic sketch showing the overview of the \emph{motion segmentation} algorithm described in~\cite{mahalingam2023human} for pouring the contents from a $\id{soup\_can}$ into a $\id{bowl}$. (\textbf{b}) Simply transferring the constant screw segments (shown using dashed lines) extracted from the demonstration of pouring using a $\id{soup\_can}$ to the execution of pouring using a $\id{box}$ results into a failure as the objects are colliding with each other.}
\end{figure*}

\subsection{Manipulation Planning from a Single Kinesthetic Demonstration with Different Objects }

A kinesthetic demonstration of a task, given by holding the hand of the robot in a zero-gravity mode, traces an end-effector path that satisfies the motion constraints \footnote{Task constraints, such as ``\emph{maintaining the upright pose}'' of a liquid-filled cup before pouring the liquid out of it into a bowl, are implicitly encoded in the end-effector path of the demonstration.} of the task. Let $\mathcal{J} \subset \mathbb{R}^l$ be the \emph{joint space}, that is, the set of all possible joint configurations of the robot, where $l$ is the number of joints of the robot's manipulator arm. For a manipulation task instance, a kinesthetic demonstration, $\bm{\Theta}$, is recorded as \emph{a discrete sequence of points from a joint space path}, i.e., a sequence of joint angle configurations $\bm{\Theta} = \left\langle \bm{\theta}^{(1)}, \cdots, \bm{\theta}^{(p)} \right\rangle$, where $\bm{\theta}^{(j)} \in \mathcal{J}$, $j = 1, 2, \cdots, p$ is a vector of dimension $l$, whose $i$-th component, $\theta_i^{(j)}$, represents the $i$-th joint angle of the robot arm.

Let $\mathcal{D} = \left(t_d, \relframe{{\bm{\Theta}}}{}{d}{}{}\right)$ be the demonstration of a task, where $t_d = \left(\relframe{\mathbf{g}}{}{d}{}{B_r}, \relframe{\mathcal{O}}{}{d}{}{r}, \relframe{\mathbf{g}}{}{d}{}{B_s}, \relframe{\mathcal{O}}{}{d}{}{s}\right)$ is the demonstrated task instance containing the pose and geometric information of the objects used in the task and $\relframe{{\bm{\Theta}}}{}{d}{}{} = \left\langle \relframe{{\bm{\theta}}}{}{d}{(1)}{}, \cdots, \relframe{{\bm{\theta}}}{}{d}{(p)}{}\right\rangle$ is the demonstrated motion, where $\relframe{{\bm{\theta}}}{}{d}{(i)}{} \in \mathbb{R}^l$. Let $t_n = \left(\relframe{\mathbf{g}}{}{n}{}{B_r}, \relframe{\mathcal{O}}{}{n}{}{r}, \relframe{\mathbf{g}}{}{n}{}{B_s}, \relframe{\mathcal{O}}{}{n}{}{s}\right)$ be a new instance of the same task containing two objects that may have different poses and/or geometries from those used in $t_d$. Let $\beta_r (\bm{\theta}) \subset \mathbb{R}^3$ be the set of points occupied by the new primary object held in the robot's hand expressed in the world frame. Since the primary object moves as the robot configuration, $\bm{\theta}$ changes, $\beta_r$ is a function of $\bm{\theta}$. Let $\beta_s \subset \mathbb{R}^3$ be the set of points occupied by the new passive object in the world frame. As the passive object is stationary, $\beta_s$ does not change. The problem of manipulation planning from a single kinesthetic demonstration using objects that are different from those used in the demonstration is as follows:\\
\emph{Given a demonstration $\mathcal{D}$ and a new task instance $t_n$ as defined above, compute a manipulation plan, i.e., a sequence of joint angles $\relframe{{\bm{\Theta}}}{}{n}{}{} = \left\langle \relframe{{\bm{\theta}}}{}{n}{(1)}{}, \cdots, \relframe{{\bm{\theta}}}{}{n}{(q)}{} \right\rangle$, where $\relframe{{\bm{\theta}}}{}{n}{(i)}{} \in \mathbb{R}^l$ is the vector of joint angles of a $l$-DoF manipulator, such that}
\begin{enumerate*}[label=(\roman*)]
    \item \emph{the motion constraints that characterize the manipulation task are satisfied,}

    \item \emph{the primary object does not collide with the passive object, i.e., $\beta_r \left(\relframe{{\bm{\theta}}}{}{n}{(i)}{}\right) \cap \beta_s = \varnothing, \forall i$, and}

    \item \emph{the task is successfully completed, i.e., a function of the form $f\left(\relframe{{\bm{\theta}}}{}{n}{(i)}{}, \relframe{\mathcal{O}}{}{n}{}{r}, \relframe{\mathcal{O}}{}{n}{}{s}\right) \leq 0$ is satisfied $\forall i \geq i_0$, with $0 \leq i_0 \leq q$.}
\end{enumerate*}

\ignore{
Given the following:
\begin{itemize}
    \item A demonstration $\left(t_d, \relframe{{\bm{\Theta}}}{}{d}{}{}\right)$ of a task, where $t_d = \left(\relframe{\mathbf{g}}{}{d}{}{B_r}, \relframe{\mathcal{O}}{}{d}{}{r}, \relframe{\mathbf{g}}{}{d}{}{B_s}, \relframe{\mathcal{O}}{}{d}{}{s}\right)$ is the demonstrated task instance containing the pose and geometric information of the objects used in the task and $\relframe{{\bm{\Theta}}}{}{d}{}{} = \left\langle \relframe{{\bm{\theta}}}{}{d}{(1)}{}, \cdots, \relframe{{\bm{\theta}}}{}{d}{(p)}{}\right\rangle$ is the demonstrated motion, where $\relframe{{\bm{\theta}}}{}{d}{(i)}{} \in \mathbb{R}^l$.

    \item A new instance $t_n = \left(\relframe{\mathbf{g}}{}{n}{}{B_r}, \relframe{\mathcal{O}}{}{n}{}{r}, \relframe{\mathbf{g}}{}{n}{}{B_s}, \relframe{\mathcal{O}}{}{n}{}{s}\right)$ of the same task containing two objects that may have a pose and/or geometry different from those used in $t_d$.
\end{itemize}

The objective is to compute a manipulation plan for $t_n$, i.e. a sequence of joint angles $\relframe{{\bm{\Theta}}}{}{n}{}{} = \left\langle \relframe{{\bm{\theta}}}{}{n}{(1)}{}, \cdots, \relframe{{\bm{\theta}}}{}{n}{(q)}{} \right\rangle$, where $\relframe{{\bm{\theta}}}{}{n}{(i)}{} \in \mathbb{R}^l$ is the vector of joint angles of a $l$-DoF manipulator, such that the motion constraints that characterize the manipulation task are satisfied. We also require that the primary object does not collide with the passive object and that the task is performed appropriately. 
}

The exact form of the function $f(\cdot)$ is task-dependent. For example, in pouring, the goal is to transfer the secondary object from the primary object (that holds the contents) to the passive object (the receiving container) without spills. Therefore, the function $f(\cdot)$ should reflect this physical constraint (see Section~\ref{sec:motion-transfer}, for a definition of $f$ for pouring). Note that ensuring that motion constraints are satisfied does not ensure that the content will be poured inside the receiving container when the dimensions of the pouring and receiving containers change.

\section{Motion Transfer from Demonstration to New Task Instance}
\label{sec:motion-transfer}
We now describe our solution approach for generalizing from a single kinesthetic demonstration to task instances with different objects. Our approach is based on a screw-geometric representation of the demonstration as presented in~\cite{mahalingam2023human}. Therefore, to make this paper self-contained, we will first summarize the approach presented in~\cite{mahalingam2023human} of generalizing from a single demonstration to different task instances, where the objects remain the same, but the poses of the primary and passive objects change. We will then discuss the limitations of the approach in~\cite{mahalingam2023human} for plan generation with different object geometries and present our approach to overcoming these limitations. 

\ignore{
We now describe the core idea of the generation of a manipulation plan for a queried task instance $t_n$ using a chosen demonstration $\left(t_d, \relframe{{\bm{\Theta}}}{}{d}{}{}\right)$ performed on functionally similar objects. For simplicity, we assume that the task involves a single passive object.

Let the new task instance $t_n$ be $\left(\relframe{\mathbf{g}}{}{n}{}{B_r}, \relframe{\mathcal{O}}{}{n}{}{r}, \relframe{\mathbf{g}}{}{n}{}{B_s}, \relframe{\mathcal{O}}{}{n}{}{s}\right)$ and the selected demonstration be $\left(t_d, \relframe{{\bm{\Theta}}}{}{d}{}{}\right)$, where the demonstrated task instance $t_d = \left(\relframe{\mathbf{g}}{}{d}{}{B_r}, \relframe{\mathcal{O}}{}{d}{}{r}, \relframe{\mathbf{g}}{}{d}{}{B_s}, \relframe{\mathcal{O}}{}{d}{}{s}\right)$ and $\relframe{{\bm{\Theta}}}{}{d}{}{} = \left\langle \relframe{{\bm{\theta}}}{}{d}{(1)}{}, \cdots, \relframe{{\bm{\theta}}}{}{d}{(p)}{} \right\rangle$ is the recorded sequence of joint configurations, with $\relframe{{\bm{\theta}}}{}{d}{(i)}{} \in \mathbb{R}^l$ being a vector of joint angles of a $l$-DoF manipulator.
}

\subsection{Screw Geometric Approach to Planning from a Single Demonstration}
\label{sec:demonstration_representation}
The motion planner in~\cite{mahalingam2023human} generates a motion plan for a new task instance $t_n$ that differs from the demonstrated task instance $t_d$, using the following key steps:
\begin{enumerate*}[label=(\roman*)]
    \item Decompose the demonstrated task space path of the end effector of the robot as a sequence of constant screw segments (one-parameter subgroups of $SE(3)$). These constant screw segments are represented as the sequence of \emph{guiding poses} of the end effector,  $\relframe{{\bm{\Gamma}}}{}{d}{}{}$, expressed with respect to the pose of the base frame, $\relframe{\mathbf{g}}{}{d}{}{B_s}$, of the passive object (see Fig.~\ref{fig:motion_segmentation}). 
    \item Transfer the guiding poses $\relframe{{\bm{\Gamma}}}{}{d}{}{}$ extracted from the demonstration, to a new sequence of guiding poses $\relframe{{\bm{\Gamma}}}{}{n}{}{}$, in the base frame \mbox{$\{B_s\}$} of passive object, $\relframe{\mathbf{g}}{}{n}{}{B_s}$, of the new task instance $t_n$.

    \item Use screw linear interpolation (ScLERP) and Jacobian pseudo-inverse~\cite{sarker2020screw} to generate the joint-space plan $\relframe{{\bm{\Theta}}}{}{n}{}{}$ between two consecutive guiding poses in $\relframe{{\bm{\Gamma}}}{}{n}{}{}$.
\end{enumerate*}

In step (i), to extract the motion constraints as a sequence of constant screw segments~\cite{mahalingam2023human}, we use position forward kinematics~\cite{murray2017mathematical}, in each arm configuration $\relframe{{\bm{\theta}}}{}{d}{(i)}{}$, to compute the pose of the robot's end-effector frame $\relframe{\mathbf{g}}{}{d}{(i)}{E} \in SE(3)$. Thus, the demonstration can be represented as a sequence of poses $\relframe{{\mathcal{G}}}{}{d}{}{} = \left\langle \relframe{\mathbf{g}}{}{d}{(1)}{E}, \cdots, \relframe{\mathbf{g}}{}{d}{(m)}{E} \right\rangle$ in $SE(3)$. Following~\cite{mahalingam2023human}, the path $\relframe{{\mathcal{G}}}{}{d}{}{}$ can be decomposed into a sequence of guiding poses $\relframe{{\bm{\Gamma}}}{}{d}{}{} = \left\langle \relframe{\mathbf{g}}{}{d}{(i_1)}{B_sE}, \cdots, \relframe{\mathbf{g}}{}{d}{(i_k)}{B_sE} \right\rangle$ expressed with respect to the base frame pose $\relframe{\mathbf{g}}{}{d}{}{B_s}$ of the passive object (see Fig.~\ref{fig:motion_segmentation}). For example, in the case of pouring liquid into a bowl, the sequence of poses $\relframe{{\bm{\Gamma}}}{}{d}{}{}$ can be expressed with respect to the pose of the bowl in $SE(3)$. Note that $\relframe{{\bm{\Gamma}}}{}{d}{}{}$, a subsequence of $\relframe{{\mathcal{G}}}{}{d}{}{}$, is our mathematical representation of a demonstration, and two consecutive poses $\left(\relframe{\mathbf{g}}{}{d}{(i_j)}{B_sE}, \relframe{\mathbf{g}}{}{d}{(i_{j+1})}{B_sE}\right)$ in $\relframe{{\bm{\Gamma}}}{}{d}{}{}$ represent a constant screw motion. 

In a successful demonstration, the constraints characterizing the tasks are embedded in the task space path (although there may be extraneous motions that are not task-relevant).  The constant screw segments generated in step (i) contain the task-relevant constraints implicitly present in the demonstration. The fact that constant screw segments can be used to represent path constraints has been shown in our previous work~\cite{sarker2020screw, mahalingam2023human, MahalingamPPC+24} for a multitude of tasks. Further, the constant screws provide a coordinate-free representation that does not depend on the choice of the end-effector frame (any other reference frame that has a constant rigid transformation could also have been used; we will use this fact later).  The use of ScLERP in step (iii) ensures that the constant screw constraint (i.e., the task-constraint) encoded by two consecutive guiding poses is always satisfied without explicitly enforcing it~\cite{sarker2020screw}. Therefore, the planner used in~\cite{mahalingam2023human} to generate a manipulation plan ensures that the task constraints present in the demonstration are always satisfied for the new task instance.

\begin{figure*}[!t]
    \centering
    \begin{subfigure}[t]{0.4\textwidth}
        \centering
        \includegraphics[width=\linewidth]{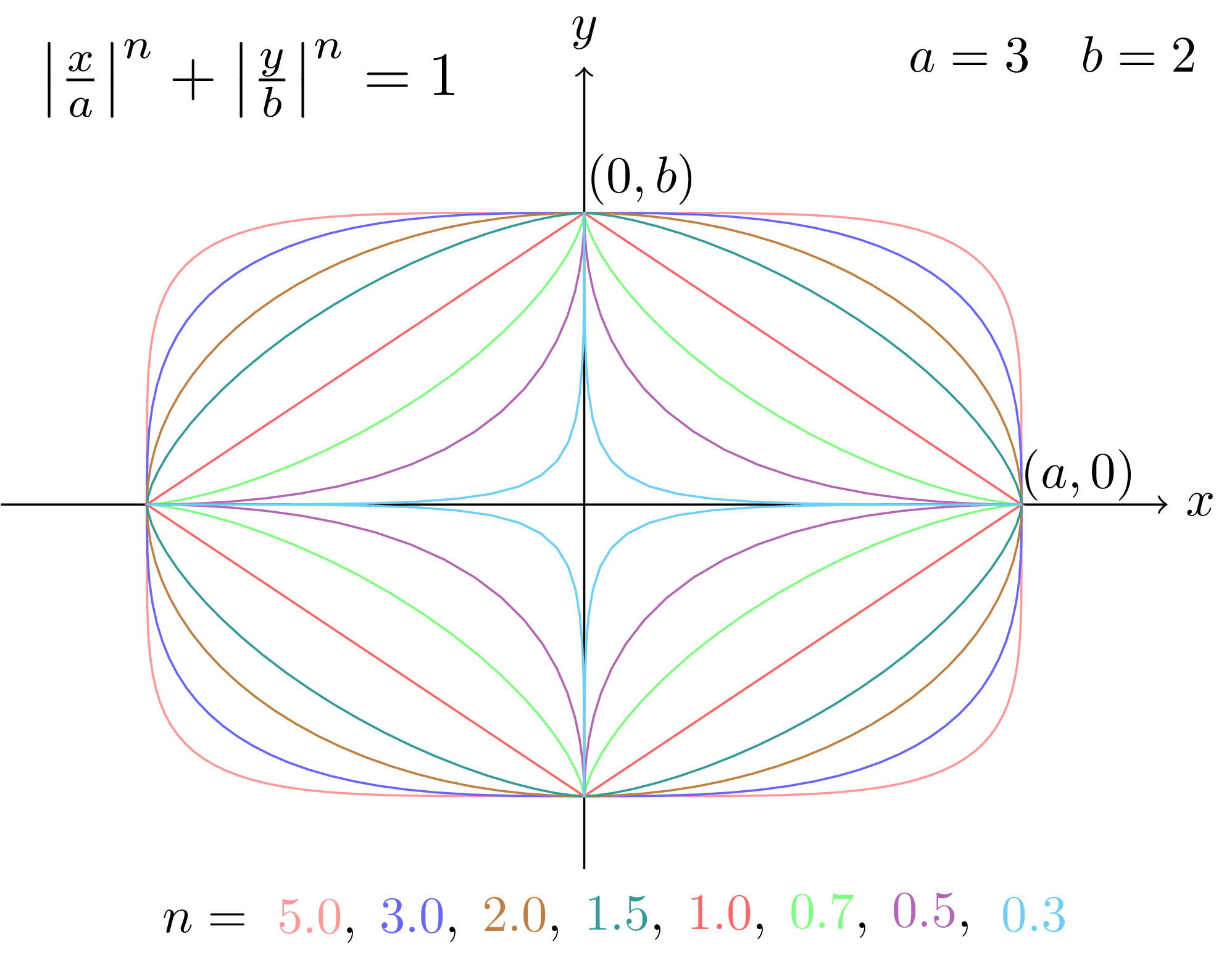}
        \caption{Geometric modeling of the opening of a container using a super-ellipse with three parameters: $a,b,$ and $n$.}
        \label{fig:super_ellipse}
    \end{subfigure}%
    ~
    \begin{subfigure}[t]{0.58\textwidth}
        \centering
        \begin{subfigure}[t]{0.5\linewidth}
            \centering
            \includegraphics[width=\linewidth]{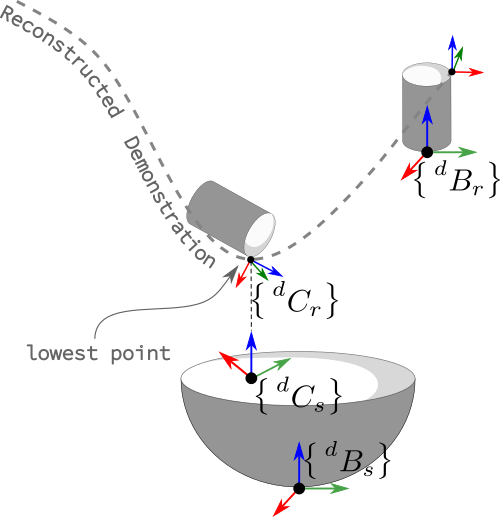}
            \caption*{(1) Demonstration}
        \end{subfigure}%
        ~
        \begin{subfigure}[t]{0.5\linewidth}
            \centering
            \includegraphics[width=\linewidth]{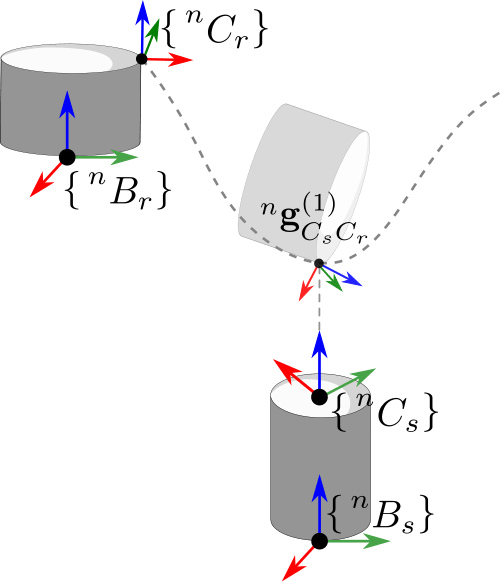}
            \caption*{(2) New task instance}
        \end{subfigure}
        \newline
        \caption{Assignment of \emph{motion-transfer} frames \mbox{$\left\{\relframe{C}{}{}{}{}\right\}$} for the primary and passive objects used in both the demonstration (left) and the new task instance (right)}
        \label{fig:c_frame_assignment}
    \end{subfigure}
    \newline\newline
    \caption{Schematic sketch of parametrization of the geometry of the rim and frame assignments.}
\end{figure*}

\mypara{Limitations in generalizing to new objects} If we use the ScLERP-based motion planner in~\cite{mahalingam2023human} to generate plans for objects different from those in the demonstration, although the plan satisfies the task motion constraints embedded in the demonstration, it may fail due to (i) collision between the primary and passive objects (see Fig.~\ref{fig:screw_extraction}) and (ii) violation of the constraint for successful task completion. This happens because the approach in~\cite{mahalingam2023human} does not consider the geometric attributes of the objects (which can be ignored if they stay the same) when transferring the screws from the demonstrated task instance $t_d$ to the new task instance $t_n$.  

\mypara{Key Technical Challenge and Our Approach} The key challenge thus becomes modifying step (ii) of transferring screws from the demonstrated task instance $t_d$ to the new task instance $t_n$, considering the geometry of the objects. 
Recall that a screw-segment is a quadruple $\left(\hat{\bm{l}}, \bm{m}, h, \theta\right)$, where the screw axis is represented in Pl\"{u}cker coordinates as $\left(\hat{\bm{l}}, \bm{m}\right)$, $\theta$ is the magnitude of rotation, and $h$ is referred to as the pitch of the screw. Thus, transferring the screw-segments w.r.t. objects of varying shapes and sizes turns out to be a problem of searching for constant screw-segments, i.e., lines in a potentially infinite space of lines, which makes the problem challenging. Our approach to solving this problem is to use an easily computable physics- and geometry-informed choice of a task-relevant reference frame for both the primary object and passive objects. We call these reference frames \emph{motion transfer frames}.


\subsection{Motion Transfer Frame}
We now discuss the choice of the motion-transfer frame using the pouring task as an example.
Two key insights while transferring a demonstration to a new instance of a complex manipulation task, involving more than one task-relevant object such as \emph{pouring, scooping, stacking etc.}, are: \begin{enumerate*}[label=(\roman*)]
  \item the motion constraints on the path of the end-effector must be maintained, and

  \item the relative motion between the \emph{critical locations} of the task-related objects must be preserved to prevent collision and ensure successful task completion.
\end{enumerate*}
For each task-relevant object we denote the \emph{motion-transfer} frame by \mbox{$\{C\}$}. The pose $\relframe{\mathbf{g}}{}{}{}{C}$ of \mbox{$\{C\}$} depends on the geometric information of the task-relevant objects and the task being performed.

\ignore{
Motion constraints are satisfied by extracting the constant screw segments from the demonstration and transferring them to the new task instance by following~\cite{mahalingam2023human}. However, simply transferring the extracted constant screw segments doesn't necessarily satisfy the other constraint of maintaining the relative motion between the \emph{critical locations} of the task-related objects (see Fig.~\ref{fig:screw_extraction}). For pouring, critical locations include where the substance leaves the primary object and where it enters the passive object.

In this work, we propose a method to address the problem of searching constant screw-segments which satisfy all the key constraints mentioned above, by introducing an additional \emph{motion-transfer} frame, \mbox{$\{C\}$} for each task-relevant object. The pose $\relframe{\mathbf{g}}{}{}{}{C}$ of \mbox{$\{C\}$} depends on the geometric information of the task-relevant objects and the task being performed.
}


The assignment of the \emph{motion-transfer} frames to the task-relevant objects requires knowledge about the set of potential locations for the origin of the frame. 
For example, for pouring, this set is usually on the lip of the container -- preferably at a notch or a sharp corner if present; for the scooping task, it's the tip of the spoon, etc. We can obtain this geometric information 
\begin{enumerate*}[label=(\roman*)]
  \item from the discrete 3D point cloud of the object, or
  \item from the 3D model of the object (e.g: CAD model) 
\end{enumerate*}
Regardless of the method, the geometric information would be a set of (discrete) points on the object to attach the \emph{motion-transfer} frame.

\subsection{Transfer of motion using the motion-transfer frame} 
\label{sec:compute_c}

The four key steps involved in the successful transfer of a demonstration to a new task instance involving objects of different shapes and sizes (geometry) are as follows:
\begin{enumerate}[label=\arabic*)]
    \item Identify the locations on the demonstrated objects critical for the task, and assign the motion-transfer frames, \mbox{$\left\{\relframe{C}{}{d}{}{r}\right\}$} and \mbox{$\left\{\relframe{C}{}{d}{}{s}\right\}$} respectively to the primary and passive objects used in the demonstration.

    \item Transfer the previously extracted \emph{guiding poses} $\relframe{{\bm{\Gamma}}}{}{d}{}{}$ (see \S\ref{sec:demonstration_representation}) to capture the pose of \mbox{$\left\{\relframe{C}{}{d}{}{r}\right\}$} represented w.r.t. the pose of \mbox{$\left\{\relframe{C}{}{d}{}{s}\right\}$}. Let's call the transferred guiding poses $\relframe{{\bm{\Gamma}}}{}{d}{}{C_sC_r}$.

    \item Identify the \emph{critical locations} on the objects used in the new task instance $t_n$, and assign their respective motion-transfer frames, \mbox{$\left\{\relframe{C}{}{n}{}{r}\right\}$} (primary) and \mbox{$\left\{\relframe{C}{}{n}{}{s}\right\}$} (passive).

    \item Finally, transfer each \emph{guiding pose} in $\relframe{{\bm{\Gamma}}}{}{d}{}{C_sC_r}$ to capture the pose of \mbox{$\left\{\relframe{C}{}{n}{}{r}\right\}$} represented w.r.t. the pose of \mbox{$\left\{\relframe{C}{}{n}{}{s}\right\}$}.
\end{enumerate}
In the following sections, we describe each step in detail with the running pouring task example. However, these steps can also be adapted to other complex tasks such as scooping, stacking objects, etc.

\subsubsection{Assigning the \mbox{$\left\{\relframe{C}{}{}{}{}\right\}$} frames to the demonstrated objects}

We model the geometry of the task-relevant objects $\relframe{\mathcal{O}}{}{d}{}{r}$, $\relframe{\mathcal{O}}{}{d}{}{s}$, $\relframe{\mathcal{O}}{}{n}{}{r}$ and $\relframe{\mathcal{O}}{}{n}{}{s}$ (containers in our example) using 4 parameters $(a, b, n, h)$, where the opening of a container can be modeled as a \emph{super-ellipse} $\left|\frac{x}{a}\right|^n + \left|\frac{y}{b}\right|^n = 1$ (see Fig.~\ref{fig:super_ellipse}) and $h$ represents its height. This gives us the flexibility to capture a wide variety of regularly shaped containers. It is also feasible to estimate these parameters~\cite{liu2022robust} from the 3D point cloud of the objects, making our modeling robust.

Fig.~\ref{fig:c_frame_assignment} shows the process of assigning the \emph{motion-transfer} frames \mbox{$\left\{\relframe{C}{}{d}{}{r}\right\}$} and \mbox{$\left\{\relframe{C}{}{d}{}{s}\right\}$} respectively to the primary and the passive objects used in the demonstration. To identify the locations on the demonstrated objects critical for the pouring task, we reconstruct the demonstrated motion using the original \emph{guiding poses} $\relframe{{\bm{\Gamma}}}{}{d}{}{}$ and the motion planner in~\cite{mahalingam2023human} (see \S\ref{sec:demonstration_representation}). The lowest point, which is the point with minimum $z$ value on the rim of the primary object used in the demonstration, along the reconstructed motion, is the identified critical point; as the substance would usually come out of this point. 
The orientation of \mbox{$\left\{\relframe{C}{}{d}{}{r}\right\}$} follows the convention of +ve $z$-axis being vertical and +ve $x$-axis along the normal to the point.

For the passive object in the demonstration, \mbox{$\left\{\relframe{C}{}{d}{}{s}\right\}$} is assigned where the primary object's lowest point vertically intersects the opening of the passive object and its orientation follows the convention of +ve $z$-axis being vertical and +ve $x$-axis pointing towards the initial pose of \mbox{$\left\{\relframe{C}{}{d}{}{r}\right\}$} on the reconstructed path.

\begin{figure}[!b]
    \centering
    \begin{subfigure}[b]{0.99\linewidth}
        \centering
        \includegraphics[width=\linewidth]{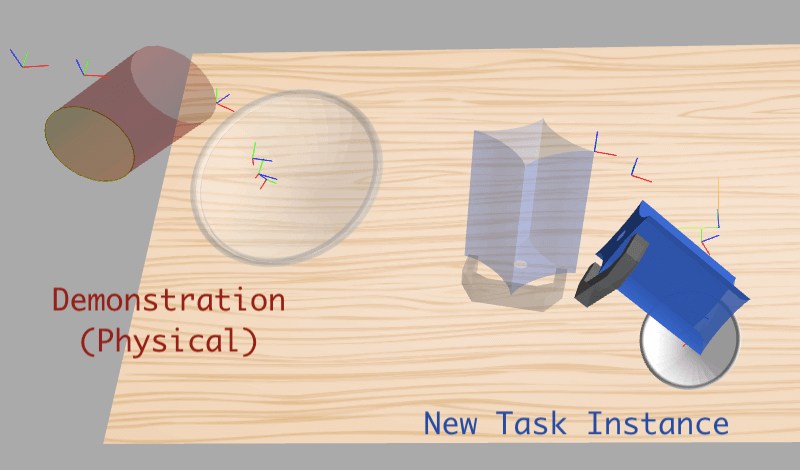}
        \captionsetup{labelformat=empty}
        \caption{$(4,4,0.7,10)$}
    \end{subfigure}

    \par\bigskip\medskip

    \begin{subfigure}[b]{0.32\linewidth}
        \centering
        \includegraphics[width=\linewidth]{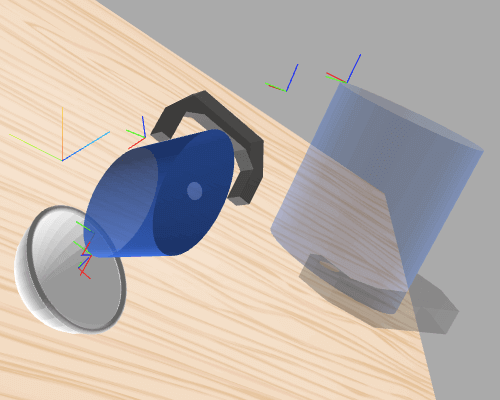}
        \captionsetup{labelformat=empty}
        \caption{$(2,4,2,8)$}
    \end{subfigure}
    \begin{subfigure}[b]{0.32\linewidth}
        \centering
        \includegraphics[width=\linewidth]{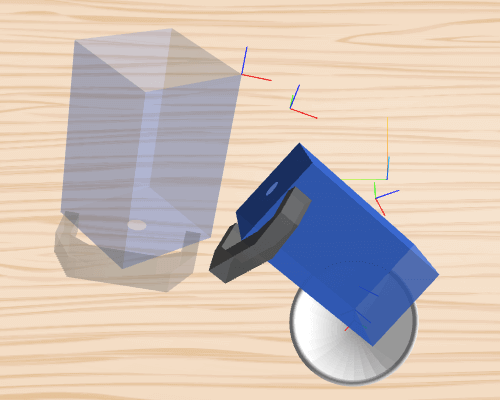}
        \captionsetup{labelformat=empty}
        \caption{$(4,4,1,10)$}
    \end{subfigure}
    \begin{subfigure}[b]{0.32\linewidth}
        \centering
        \includegraphics[width=\linewidth]{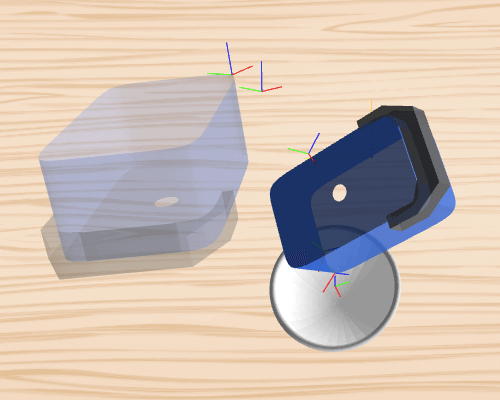}
        \captionsetup{labelformat=empty}
        \caption{$(5,3,8,4)$}
    \end{subfigure}

    \par\bigskip

    \caption{Successful \emph{pouring} task executions in simulation with virtual objects (\textcolor{blue}{blue}) of varying shapes and sizes using a \emph{single} physical demonstration (\textcolor{red}{red}). Different values of $(a,b,n,h)$ are shown for the primary object.}
    \label{fig:simulation}
    \vspace{0.5\baselineskip}
\end{figure}

\begin{table}[!b]
    \centering
    \begin{threeparttable}
        \begin{tabular}{c|c|c|c|c}
            \backslashbox[2.6cm]{Passive}{Primary} & \emph{Fat Tall} & \emph{Fat Short} & \emph{Thin Tall} & \emph{Thin Short} \\
            \midrule
            \emph{Fat Tall} & 
            \textcolor{blue}{80} vs. \textcolor{red}{34} & \textcolor{blue}{91} vs. \textcolor{red}{51} & \textcolor{blue}{93} vs. \textcolor{red}{34} & \textcolor{blue}{86} vs. \textcolor{red}{62} \\
            \emph{Fat Short} & 
            \textcolor{blue}{85} vs. \textcolor{red}{76} & \textcolor{blue}{94} vs. \textcolor{red}{78} & \textcolor{blue}{87} vs. \textcolor{red}{63} & \textcolor{blue}{93} vs. \textcolor{red}{81} \\
            \emph{Thin Tall} & 
            \textcolor{blue}{86} vs. \textcolor{red}{30} & \textcolor{blue}{69} vs. \textcolor{red}{29} & \textcolor{blue}{78} vs. \textcolor{red}{36} & \textcolor{blue}{88} vs. \textcolor{red}{43} \\
            \emph{Thin Short} & 
            \textcolor{blue}{87} vs. \textcolor{red}{65} & \textcolor{blue}{92} vs. \textcolor{red}{73} & \textcolor{blue}{89} vs. \textcolor{red}{49} & \textcolor{blue}{93} vs. \textcolor{red}{85} \\
            \bottomrule
        \end{tabular}
        \begin{tablenotes}
            \item Numbers in \textcolor{blue}{blue} and \textcolor{red}{red} respectively denote the percentage (out of 200) of successful executions with and without the \emph{motion-transfer} frames \mbox{$\left\{\relframe{C}{}{}{}{}\right\}$}
            \item \emph{Passive}: (\textbf{Thin}) $a,b < 8 \leq a,b$ (\textbf{Fat}) and (\textbf{Short}) $h < 5.5 \leq h$ (\textbf{Tall})
            \item \emph{Primary}: (\textbf{Thin}) $a,b < 3.25 \leq a,b$ (\textbf{Fat}) and (\textbf{Short}) $h < 10 \leq h$ (\textbf{Tall})
        \end{tablenotes}
        \caption{Collision-free plans with virtual objects.}
        \label{tab:simulation_result}
    \end{threeparttable}
    \vspace{\baselineskip}
\end{table}

\subsubsection{Transferring $\relframe{{\mathbf{\Gamma}}}{}{d}{}{}$ to represent the relative motion of \mbox{$\left\{\relframe{C}{}{d}{}{r}\right\}$} with respect to \mbox{$\left\{\relframe{C}{}{d}{}{s}\right\}$}}

The relative transformation of the pose $\relframe{\mathbf{g}}{}{}{}{B} \in SE(3)$ of the reference frame $\{B\}$ is expressed relative to the pose $\relframe{\mathbf{g}}{}{}{}{A} \in SE(3)$ of reference frame $\{A\}$ as $\relframe{\mathbf{g}}{}{}{}{AB} = \relframe{\mathbf{g}}{}{}{}{B}\left(\relframe{\mathbf{g}}{}{}{}{A}\right)^{-1}$.
\ignore{such that the unit dual quaternion $\relframe{\mathbf{D}}{}{}{}{AB}$ of $\relframe{\mathbf{g}}{}{}{}{AB}$ is
$$\relframe{\mathbf{D}}{}{}{}{AB} = \relframe{\mathbf{D}}{}{}{*}{A} \otimes \relframe{\mathbf{D}}{}{}{}{B}$$
where $\relframe{\mathbf{D}}{}{}{*}{A}$ is the conjugate of the unit dual quaternion of $\relframe{\mathbf{g}}{}{}{}{A}$ and $\relframe{\mathbf{D}}{}{}{}{B}$ is the unit dual quaternion of $\relframe{\mathbf{g}}{}{}{}{B}$.
} 
Once we assign the poses $\relframe{\mathbf{g}}{}{d}{}{C_r}$ and $\relframe{\mathbf{g}}{}{d}{}{C_s}$ of \mbox{$\left\{\relframe{C}{}{d}{}{r}\right\}$} and \mbox{$\left\{\relframe{C}{}{d}{}{s}\right\}$} respectively, we determine their relative homogeneous transformation by transferring the previously extracted \emph{guiding poses} $\relframe{{\bm{\Gamma}}}{}{d}{}{}$ (see \S\ref{sec:demonstration_representation}) to capture the poses of \mbox{$\left\{\relframe{C}{}{d}{}{r}\right\}$} represented w.r.t. the pose $\relframe{\mathbf{g}}{}{d}{}{C_s}$ of \mbox{$\left\{\relframe{C}{}{d}{}{s}\right\}$} and we denote the corresponding sequence as $\relframe{{\bm{\Gamma}}}{}{d}{}{C_sC_r}$, where

$\relframe{{\bm{\Gamma}}}{}{d}{}{C_sC_r} = \left\langle \relframe{\mathbf{g}}{}{d}{(1)}{C_sC_r}, \cdots, \relframe{\mathbf{g}}{}{d}{(k)}{C_sC_r} \right\rangle$, where, $\forall i\in [1,k]$
\ignore{
\begin{align}
    \relframe{\mathbf{g}}{}{d}{(i)}{C_sC_r} & = rel\left(\relframe{\mathbf{g}}{}{d}{}{C_sB_s}, \relframe{\mathbf{g}}{}{d}{(i)}{B_sC_r}\right)\quad \forall i\in [1,k] \label{eq:demo_c_frame}\\
    \relframe{\mathbf{g}}{}{d}{}{C_sB_s} & = rel\left(\relframe{\mathbf{g}}{}{d}{}{C_s}, \relframe{\mathbf{g}}{}{d}{}{B_s}\right) \nonumber\\
    \relframe{\mathbf{g}}{}{d}{(i)}{B_sC_r} & = rel\left(\relframe{\mathbf{g}}{}{d}{(i)}{B_sE}, \relframe{\mathbf{g}}{}{d}{}{EC_r}\right)\quad\forall i\in [1,k]\quad \relframe{\mathbf{g}}{}{d}{(i)}{B_sE} \in \relframe{{\bm{\Gamma}}}{}{d}{}{} \nonumber\\
    \relframe{\mathbf{g}}{}{d}{}{EC_r} & = rel\left(rel\left(\relframe{\mathbf{g}}{}{}{}{E}, \relframe{\mathbf{g}}{}{d}{}{B_r}\right), rel\left(\relframe{\mathbf{g}}{}{d}{}{B_r}, \relframe{\mathbf{g}}{}{d}{}{C_r}\right)\right) \nonumber
\end{align}
}
\begin{equation}
    \label{eq:frame_transfer}
    \relframe{\mathbf{g}}{}{d}{(i)}{C_sC_r} = \left(\relframe{\mathbf{g}}{}{d}{}{B_sC_s}\right)^{-1}\relframe{\mathbf{g}}{}{d}{(i)}{B_sE}\relframe{\mathbf{g}}{}{d}{}{EB_r}\relframe{\mathbf{g}}{}{d}{}{B_rC_r}
\end{equation}

\subsubsection{Assigning \mbox{$\left\{\relframe{C}{}{}{}{}\right\}$} frames to new task instance objects}

Assignment of the \emph{motion-transfer} frame \mbox{$\left\{\relframe{C}{}{n}{}{r}\right\}$} to the primary object in the new task instance $t_n$ requires the geometric information about its opening. 
Depending on the values of $a,b$ and $n$, we select the position of the primary object's \emph{motion transfer} frame \mbox{$\left\{\relframe{C}{}{n}{}{r}\right\}$} as follows:
\begin{enumerate*}[label=(\roman*)]
    \item If $n \neq 2$ or $a \neq b$, we assign the origin of \mbox{$\left\{\relframe{C}{}{n}{}{r}\right\}$} on the lip of the primary object at the corner nearest to the passive object's base reference frame \mbox{$\left\{\relframe{B}{}{n}{}{s}\right\}$}.

    \item Otherwise, we assign it at the intersection point between the $x$-$y$ projection of the lip of the primary object and the line joining the $x$-$y$ projections of \mbox{$\left\{\relframe{B}{}{n}{}{r}\right\}$} and \mbox{$\left\{\relframe{B}{}{n}{}{s}\right\}$}.
\end{enumerate*}
The orientation of \mbox{$\left\{\relframe{C}{}{n}{}{r}\right\}$} follows the convention of +ve $z$-axis being vertical and +ve $x$-axis pointing along the normal to the lip at that point.


For the passive object, \mbox{$\left\{\relframe{C}{}{n}{}{s}\right\}$} is assigned at the center of the opening of the passive object, and its orientation follows the convention of +ve $z$-axis being vertical and +ve $x$-axis pointing towards the  assigned pose of \mbox{$\left\{\relframe{C}{}{n}{}{r}\right\}$}.
Figure~\ref{fig:c_frame_assignment} illustrates this process with two objects shaped differently from those used in the demonstration.

\begin{figure}[!b]
    \begin{subfigure}[b]{0.49\linewidth}
        \centering
        \includegraphics[width=\textwidth]{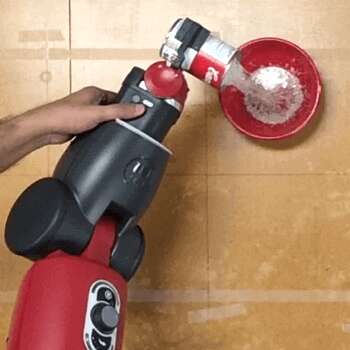}
        \captionsetup{labelformat=empty}
        \caption{\textcolor{blue}{Demo \#1} from $\id{soup\_can}$ to $\id{bowl}$}
    \end{subfigure}
    \begin{subfigure}[b]{0.49\linewidth}
        \centering
        \includegraphics[width=\textwidth]{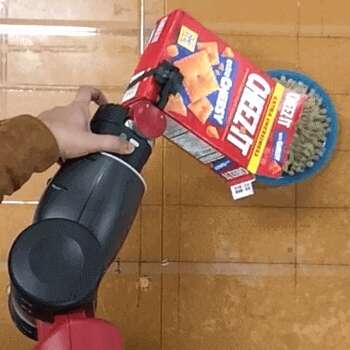}
        \captionsetup{labelformat=empty}
        \caption{\textcolor{red}{Demo \#2} from $\id{CheezIt\_box}$ to $\id{plate}$}
    \end{subfigure}
    \par\bigskip
    \caption{Two kinesthetic demonstrations of the pouring task using different \emph{primary} and \emph{passive} objects.}
    \label{fig:hardware_demonstration}
    \vspace{0.5\baselineskip}
\end{figure}

\begin{figure*}[!h]
    \begin{subfigure}[][][c]{0.5\linewidth}
        \begin{tabular}{lllll}
            \includegraphics[width=0.15\textwidth]{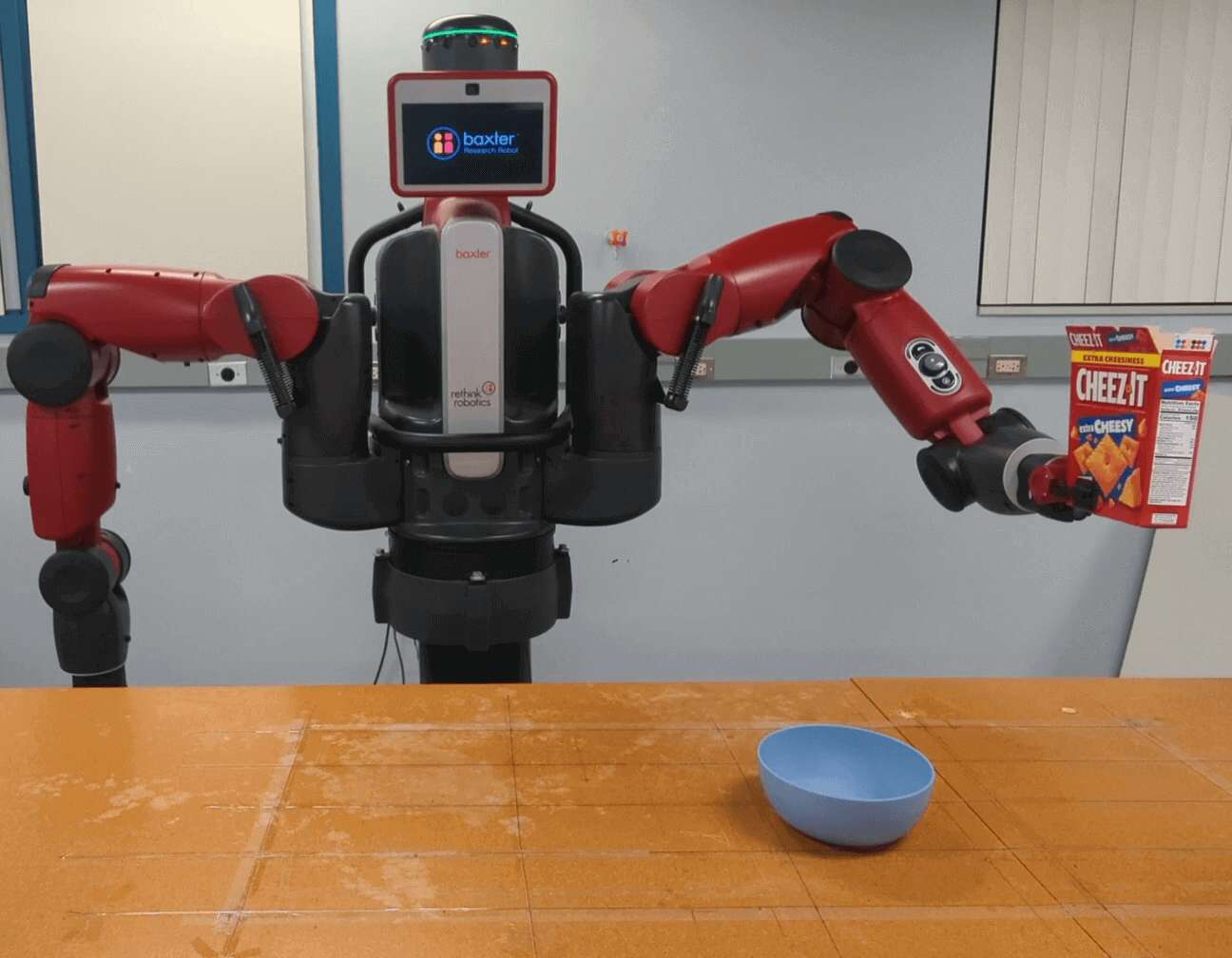} & \includegraphics[width=0.15\textwidth]{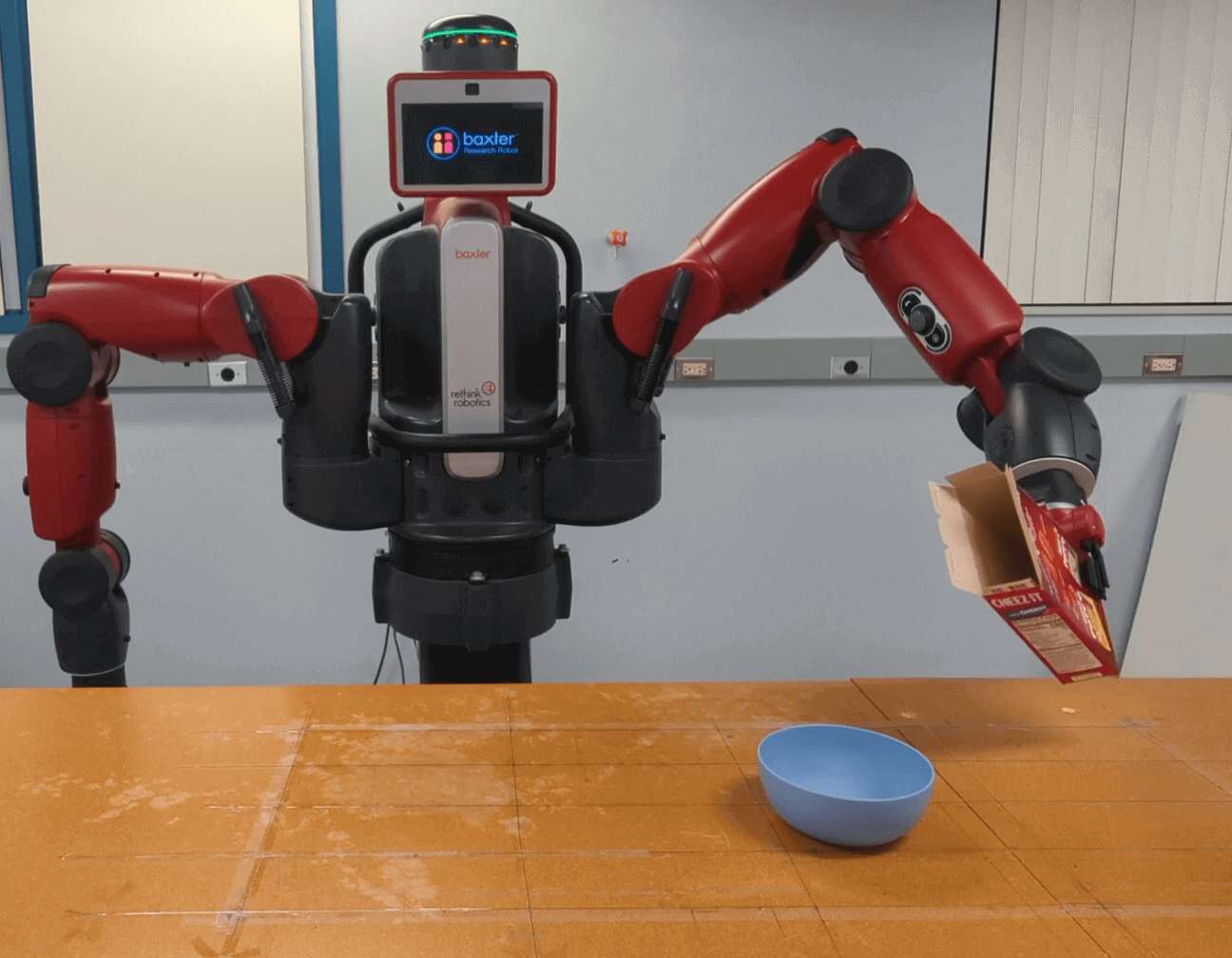} & \includegraphics[width=0.15\textwidth]{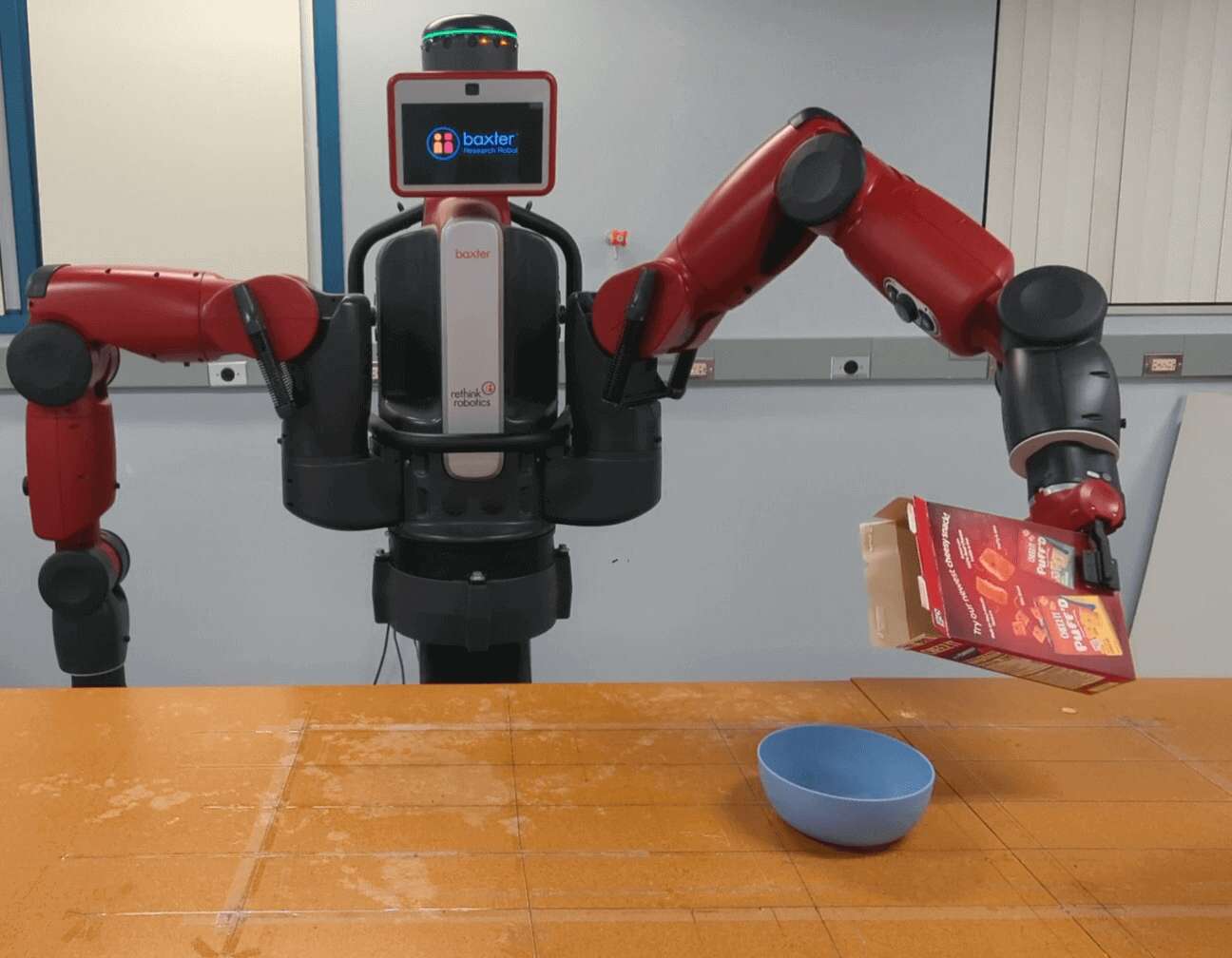} & \includegraphics[width=0.15\textwidth]{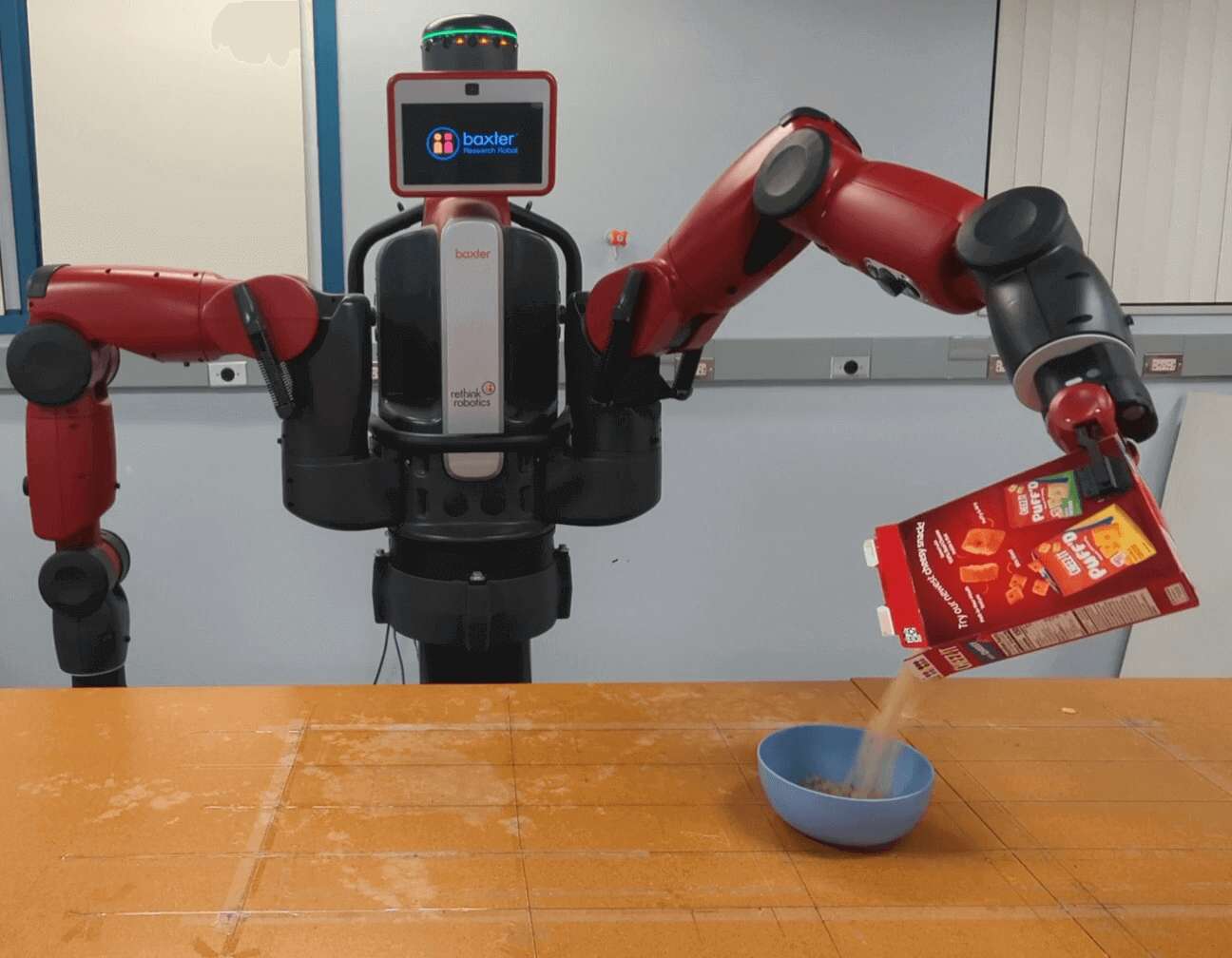} & \includegraphics[width=0.15\textwidth]{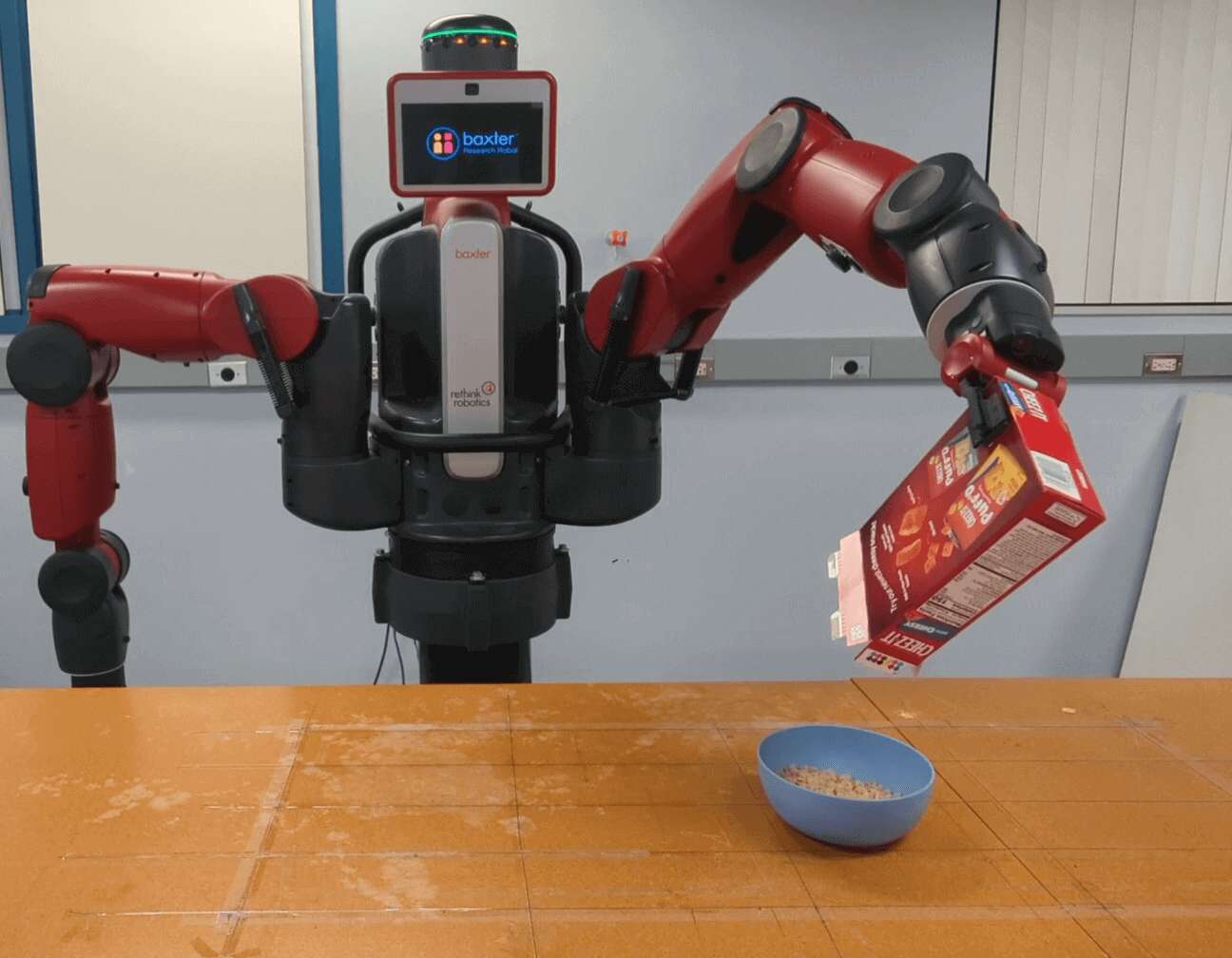}\\
            \multicolumn{5}{c}{\footnotesize{Execution from $\id{CheezIt\_box}$ to $\id{bowl}$}}\\
            \includegraphics[width=0.15\textwidth]{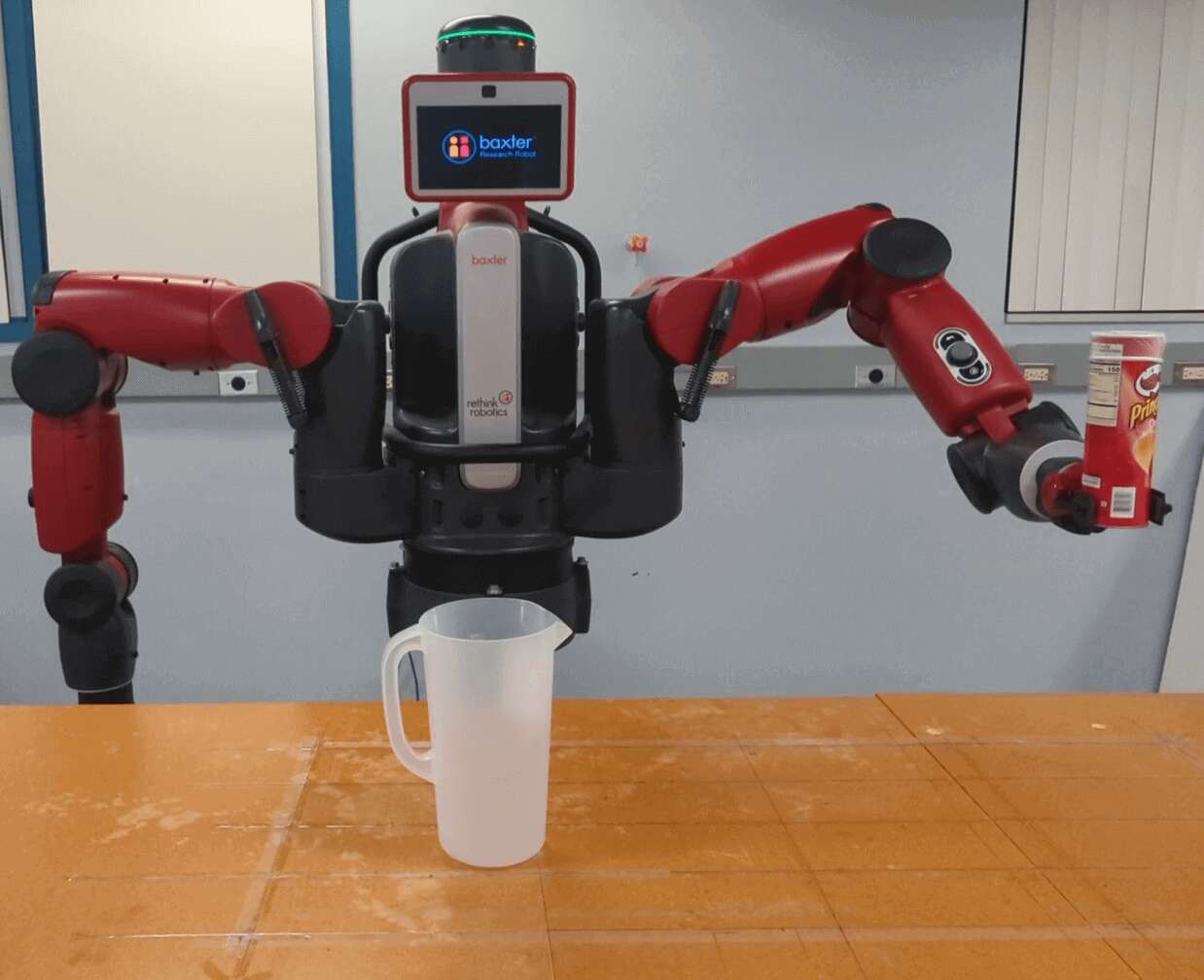} & \includegraphics[width=0.15\textwidth]{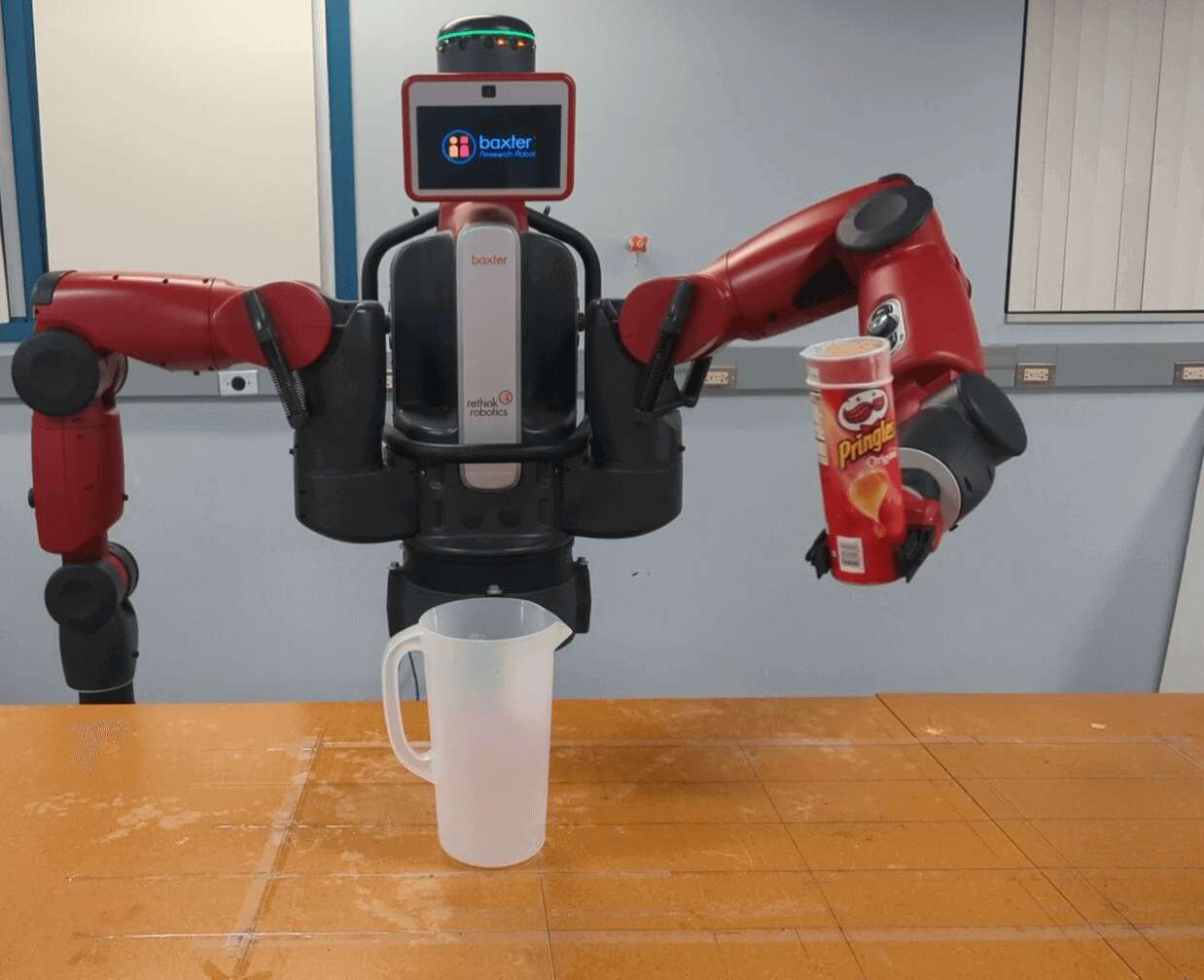} & \includegraphics[width=0.15\textwidth]{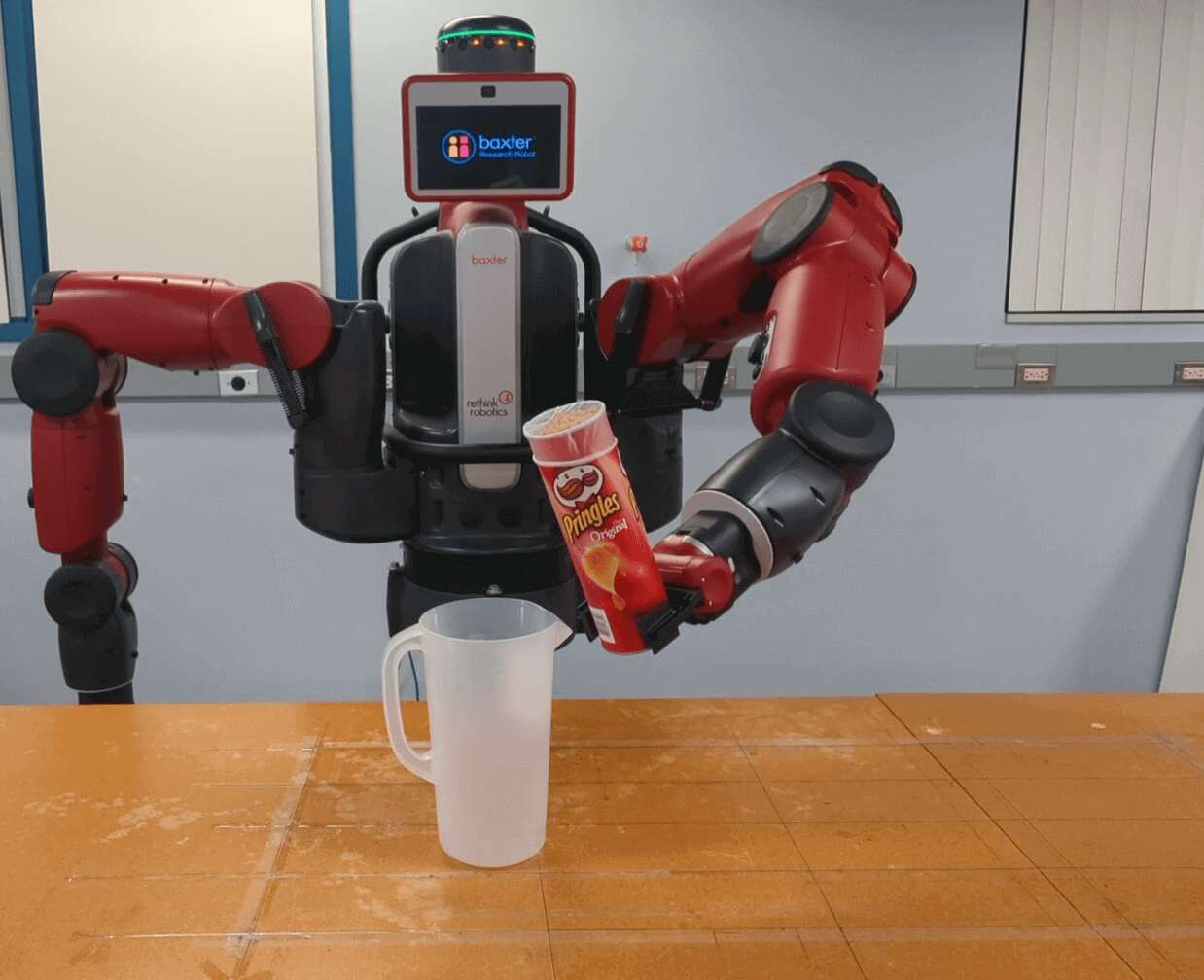} & \includegraphics[width=0.15\textwidth]{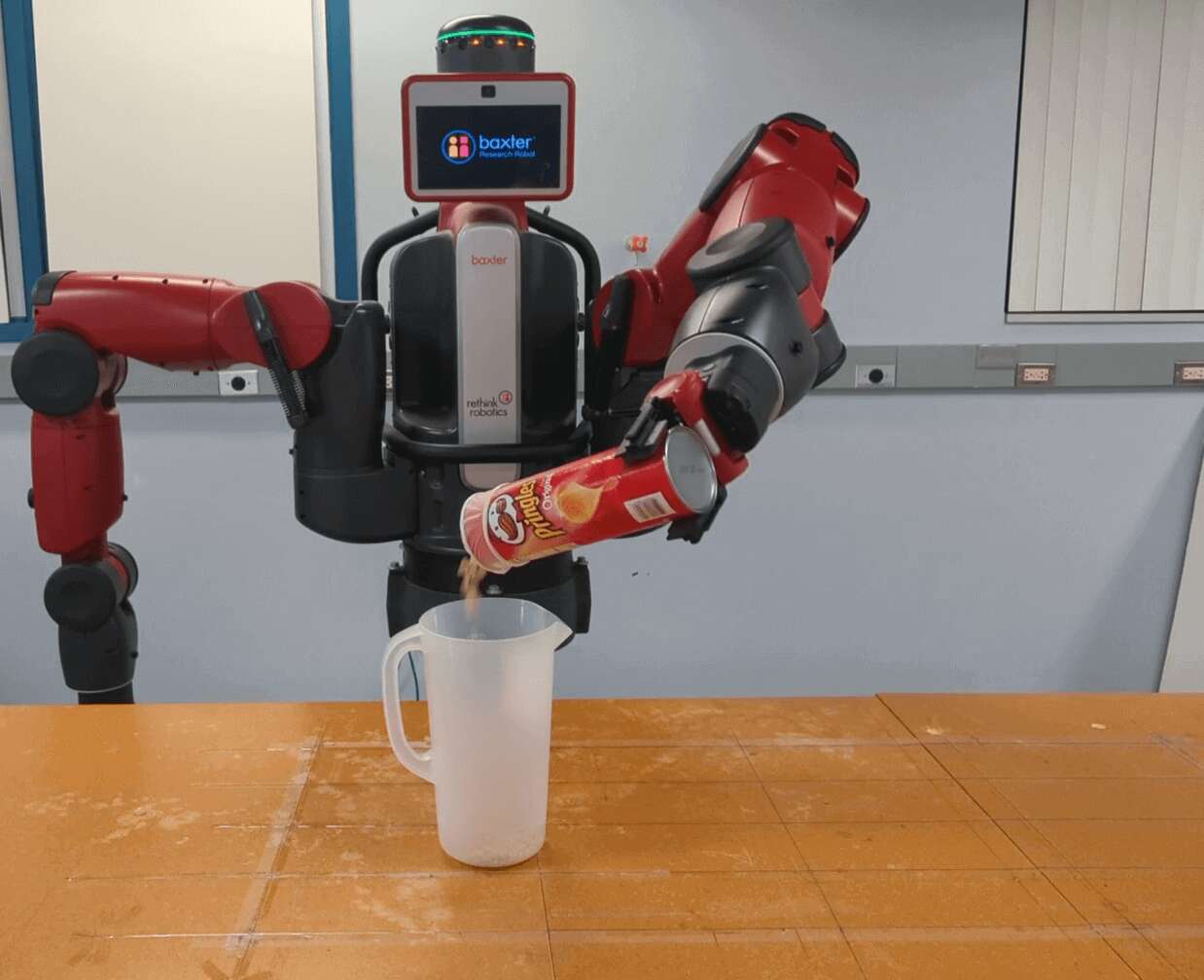} & \includegraphics[width=0.15\textwidth]{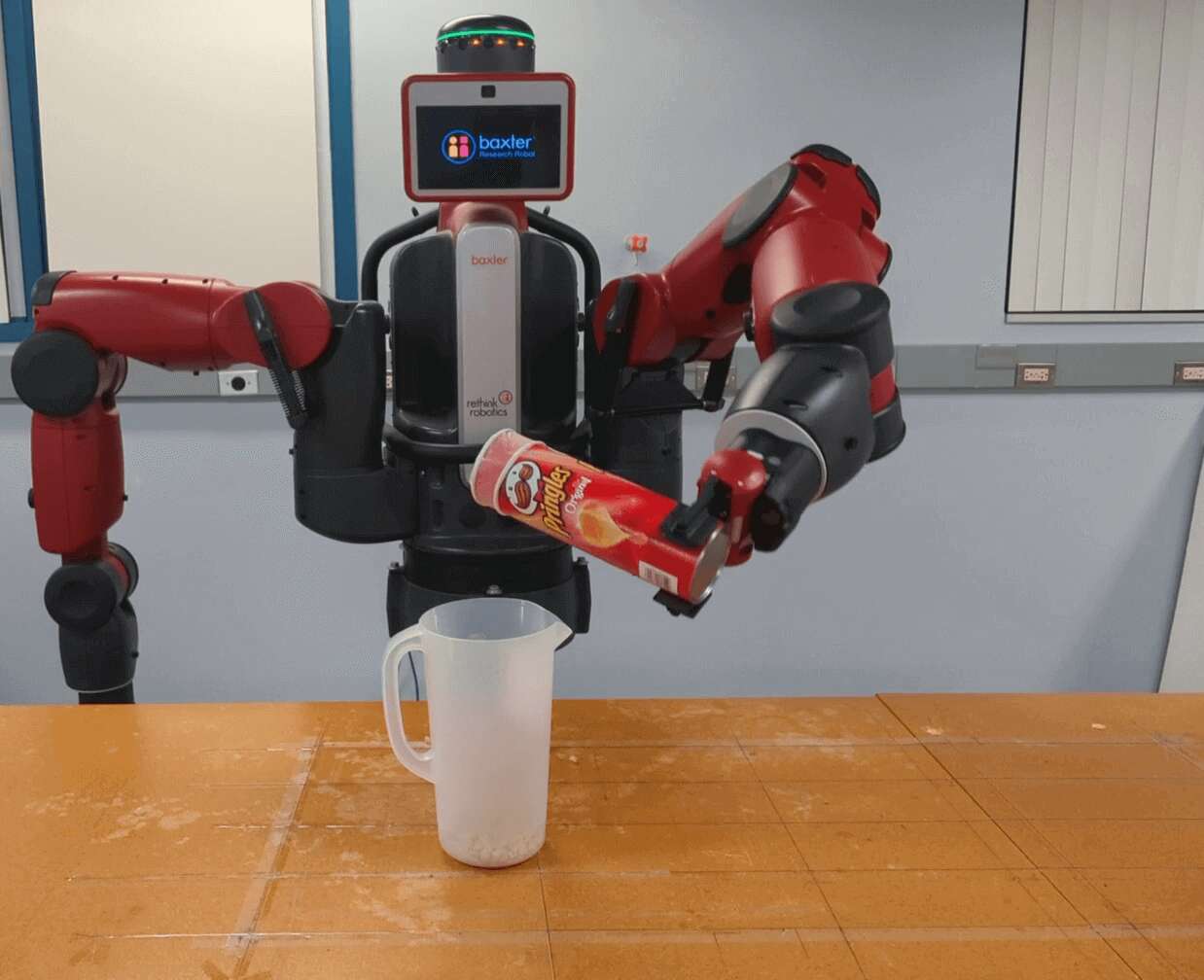}\\
            \multicolumn{5}{c}{\footnotesize{Execution from $\id{Pringles\_box}$ to $\id{pitcher}$}}\\
            \includegraphics[width=0.15\textwidth]{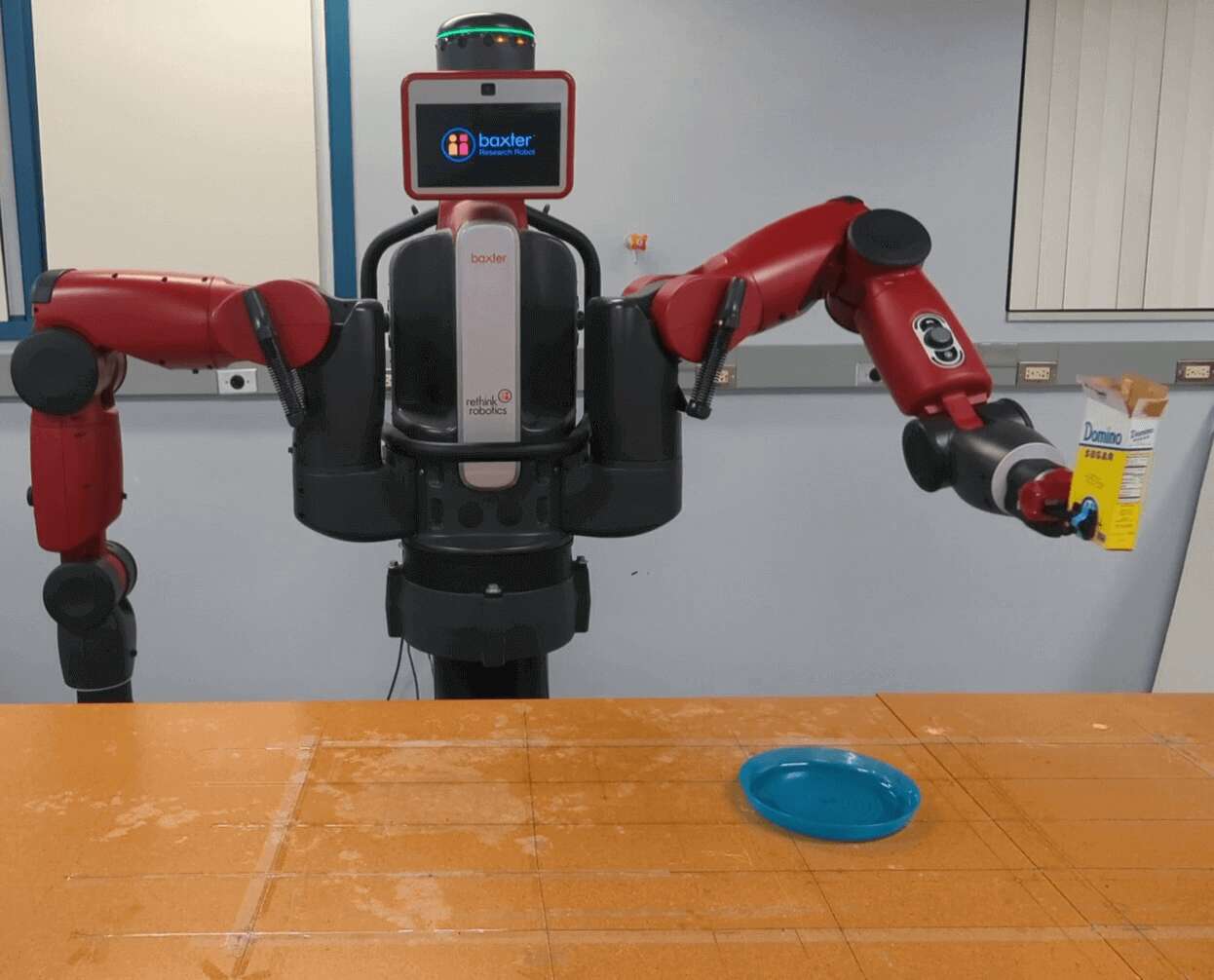} & \includegraphics[width=0.15\textwidth]{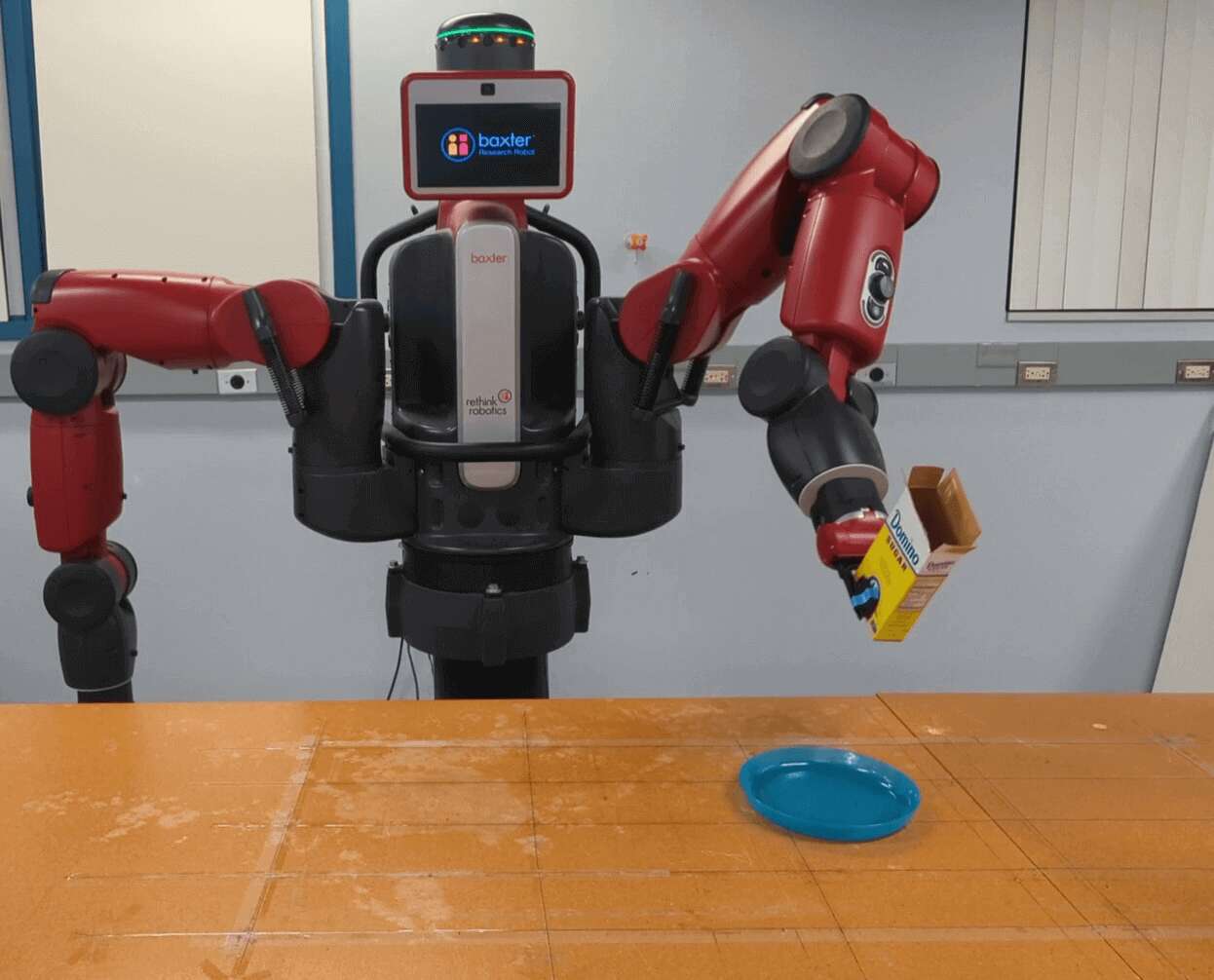} & \includegraphics[width=0.15\textwidth]{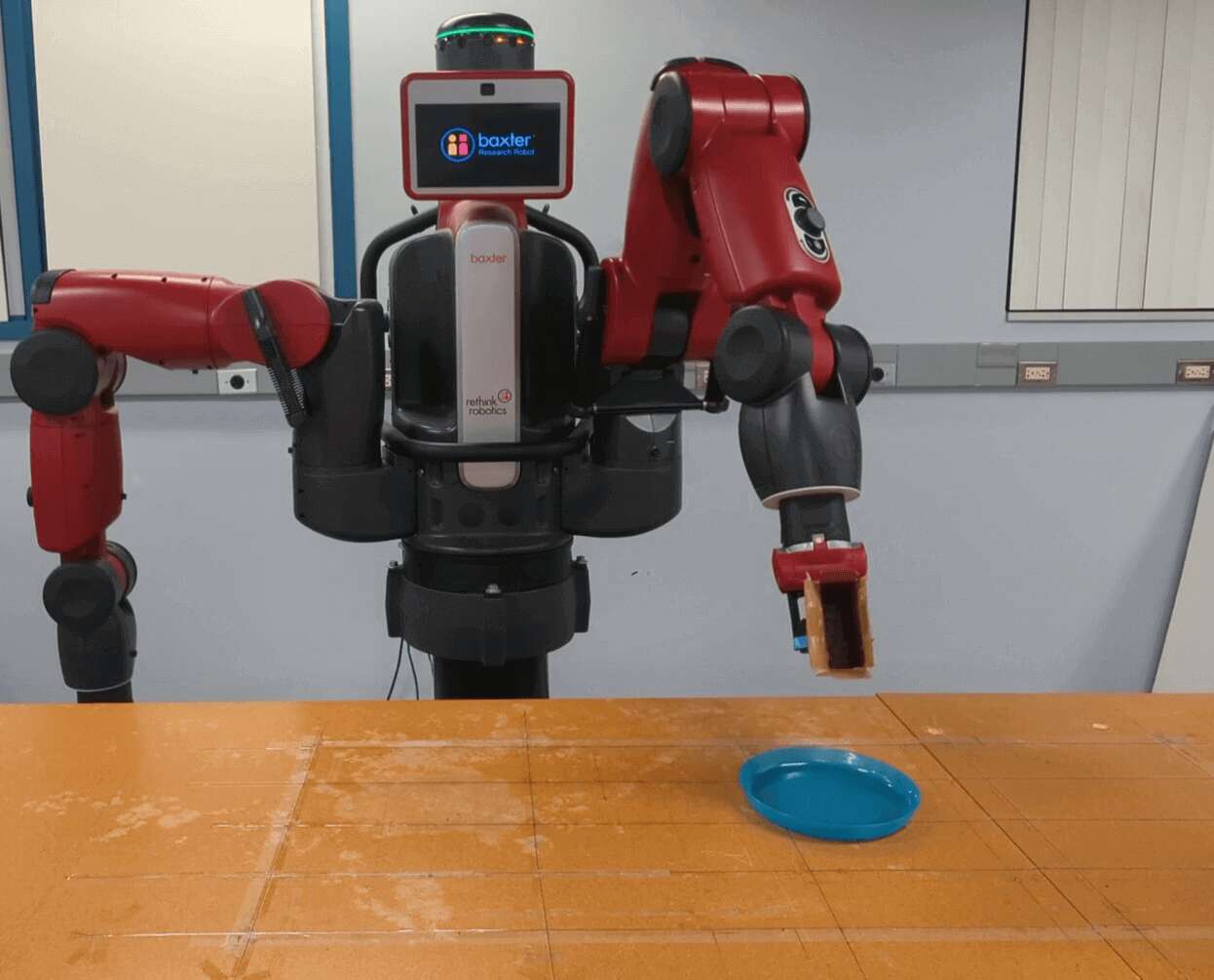} & \includegraphics[width=0.15\textwidth]{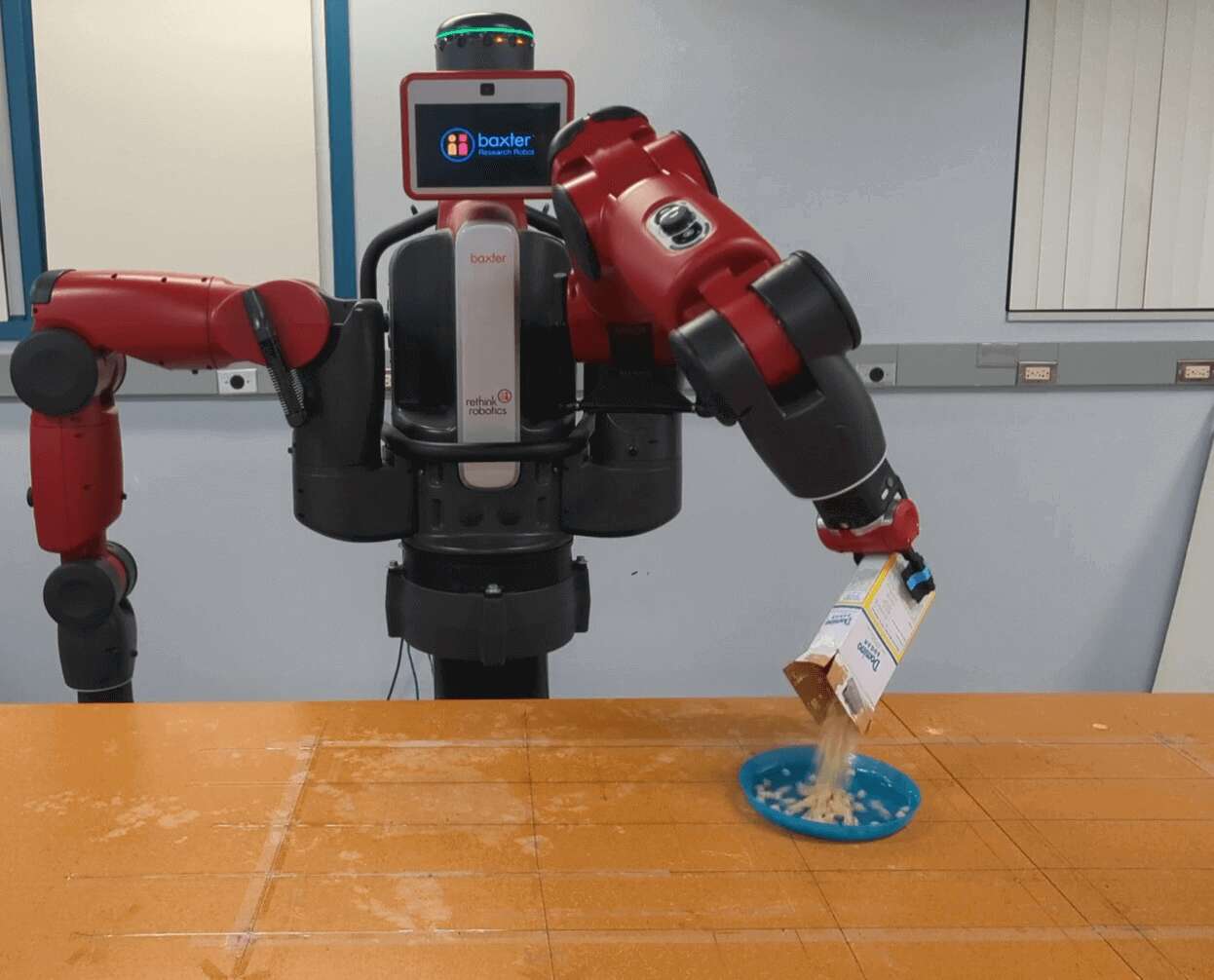} & \includegraphics[width=0.15\textwidth]{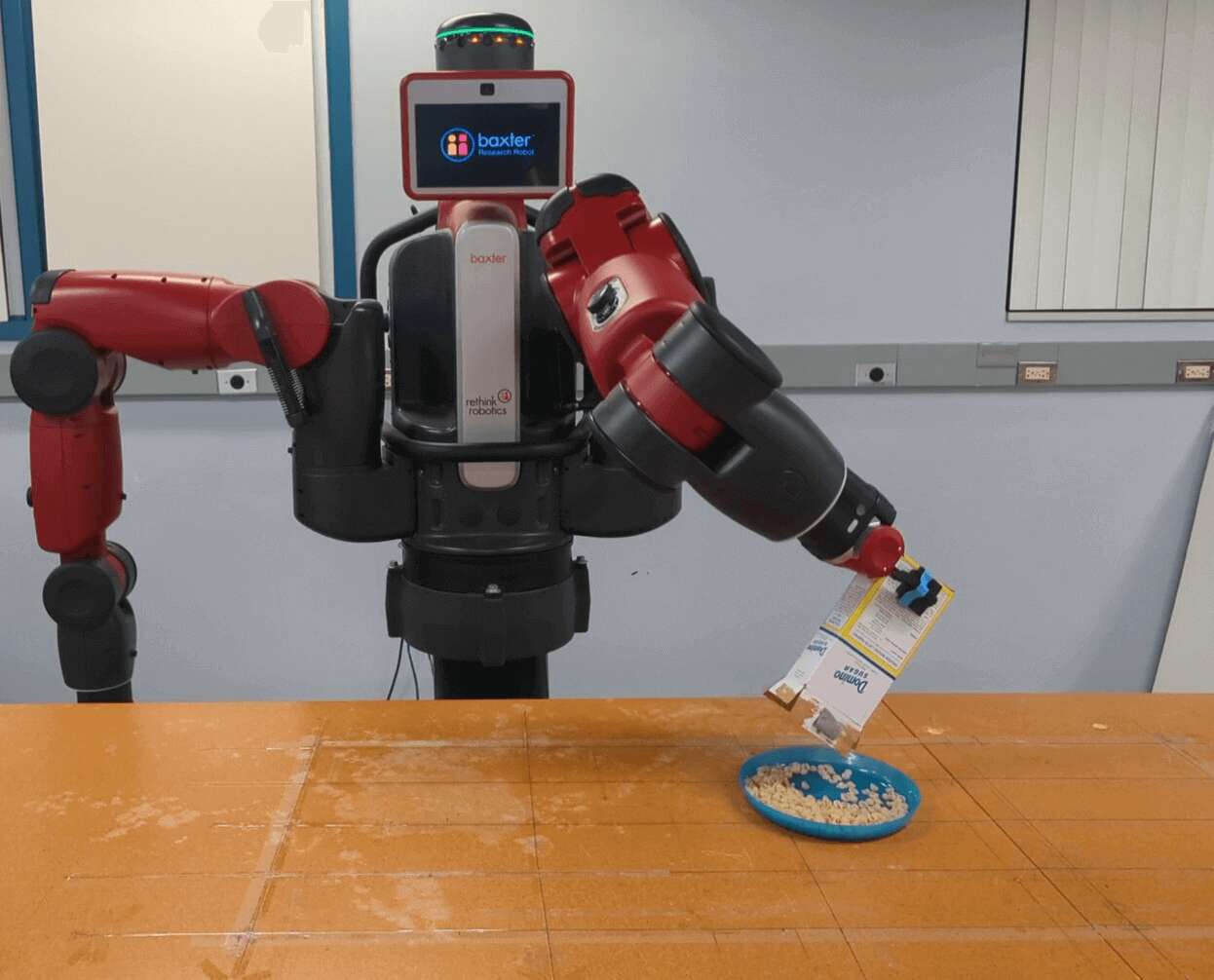}\\
            \multicolumn{5}{c}{\footnotesize{Execution from $\id{DominoSugar\_box}$ to flat $\id{plate}$}}
        \end{tabular}
        \captionsetup{labelformat=empty}
        \caption{Three task executions using \textcolor{blue}{demonstration \#1}}
    \end{subfigure}
    \begin{subfigure}[][][c]{0.5\linewidth}
        \begin{tabular}{lllll}
            \includegraphics[width=0.15\textwidth]{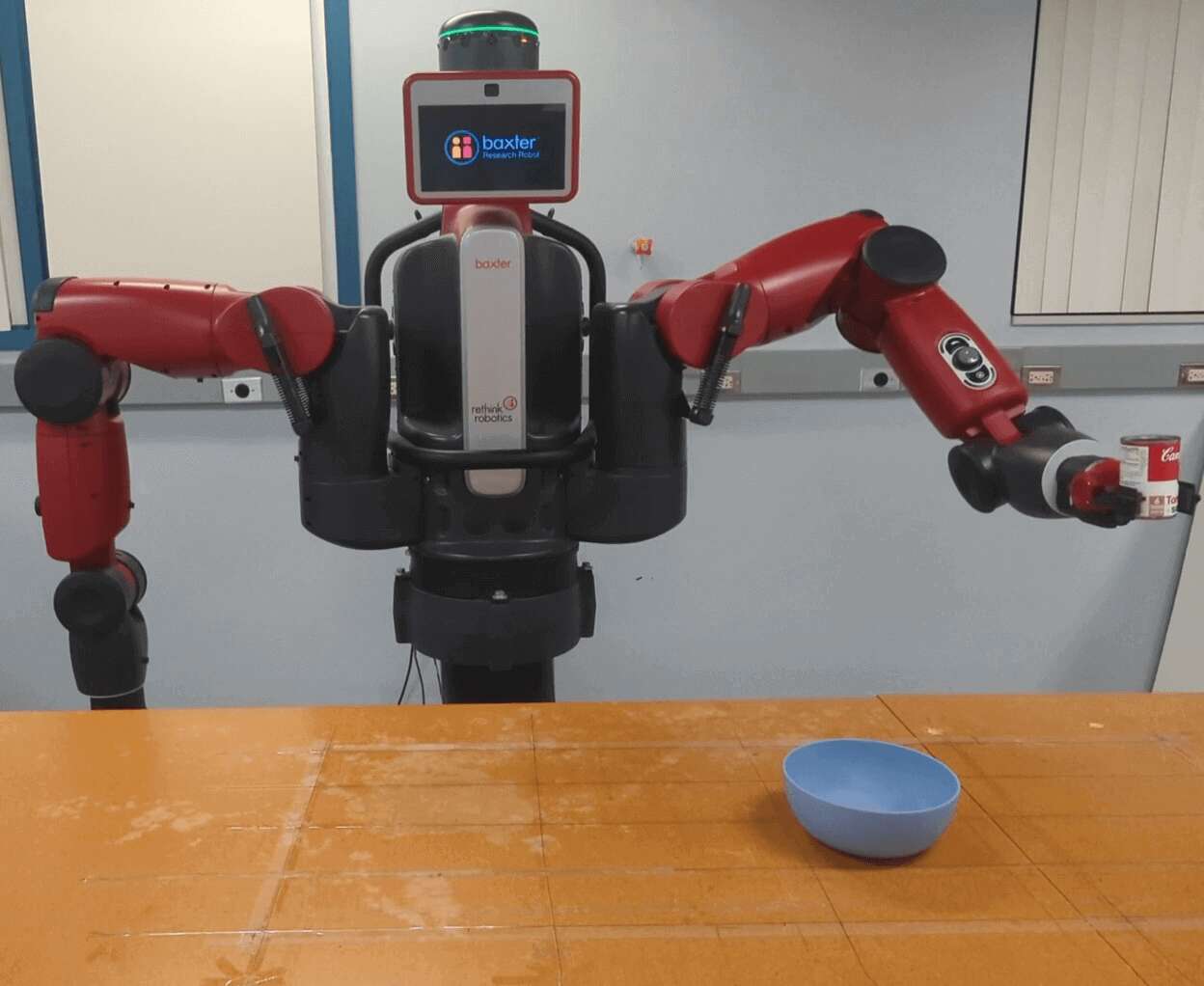} & \includegraphics[width=0.15\textwidth]{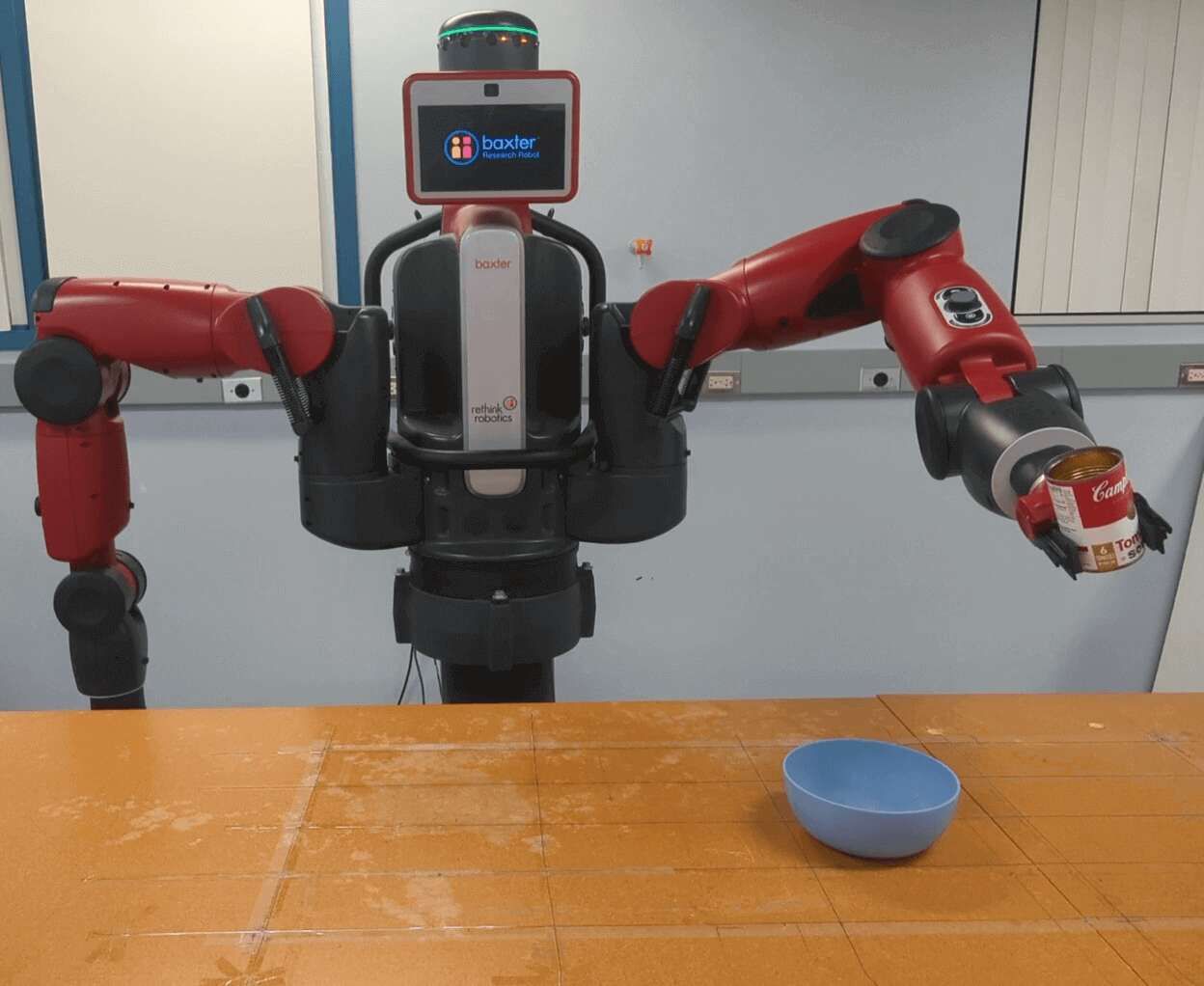} & \includegraphics[width=0.15\textwidth]{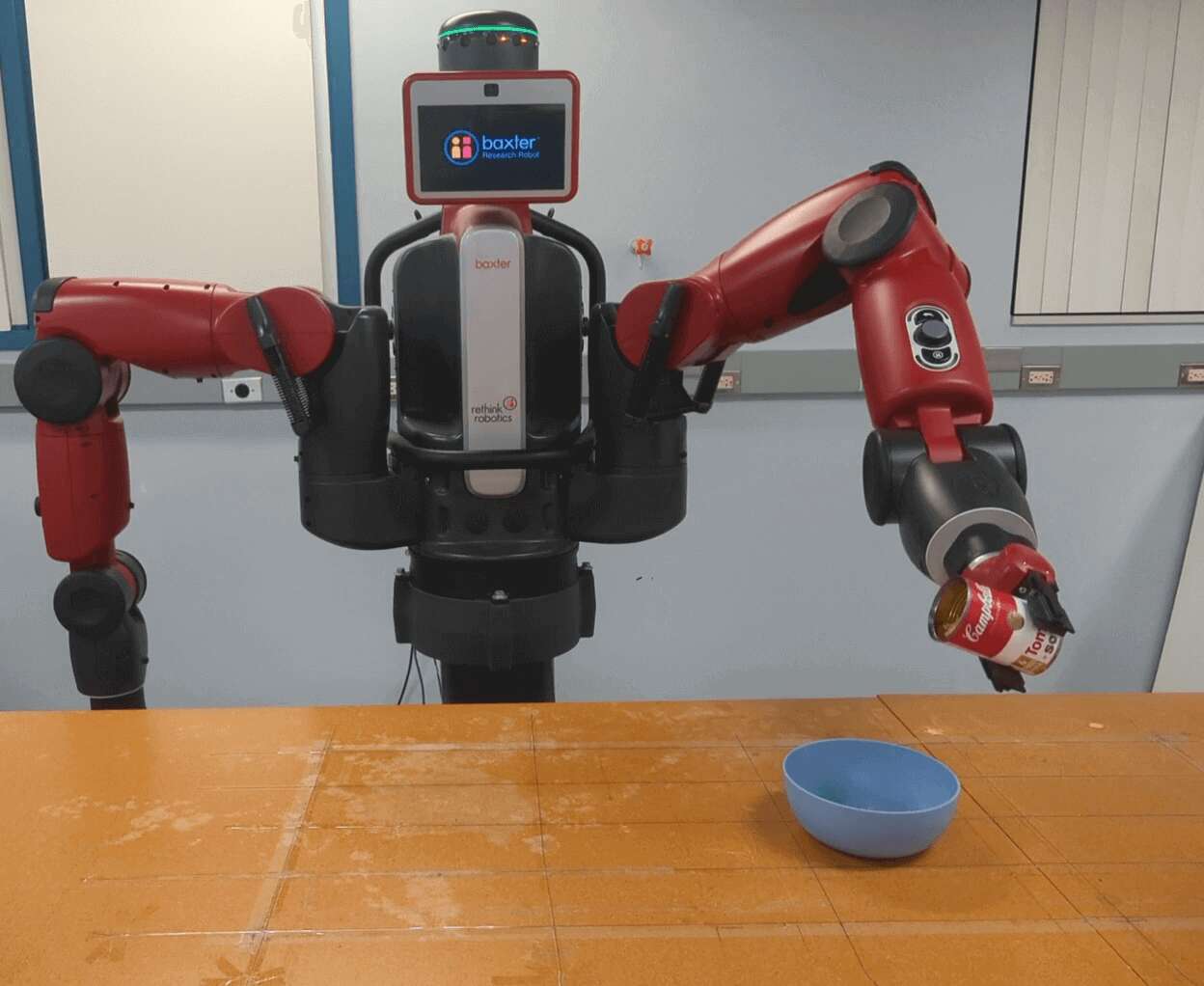} & \includegraphics[width=0.15\textwidth]{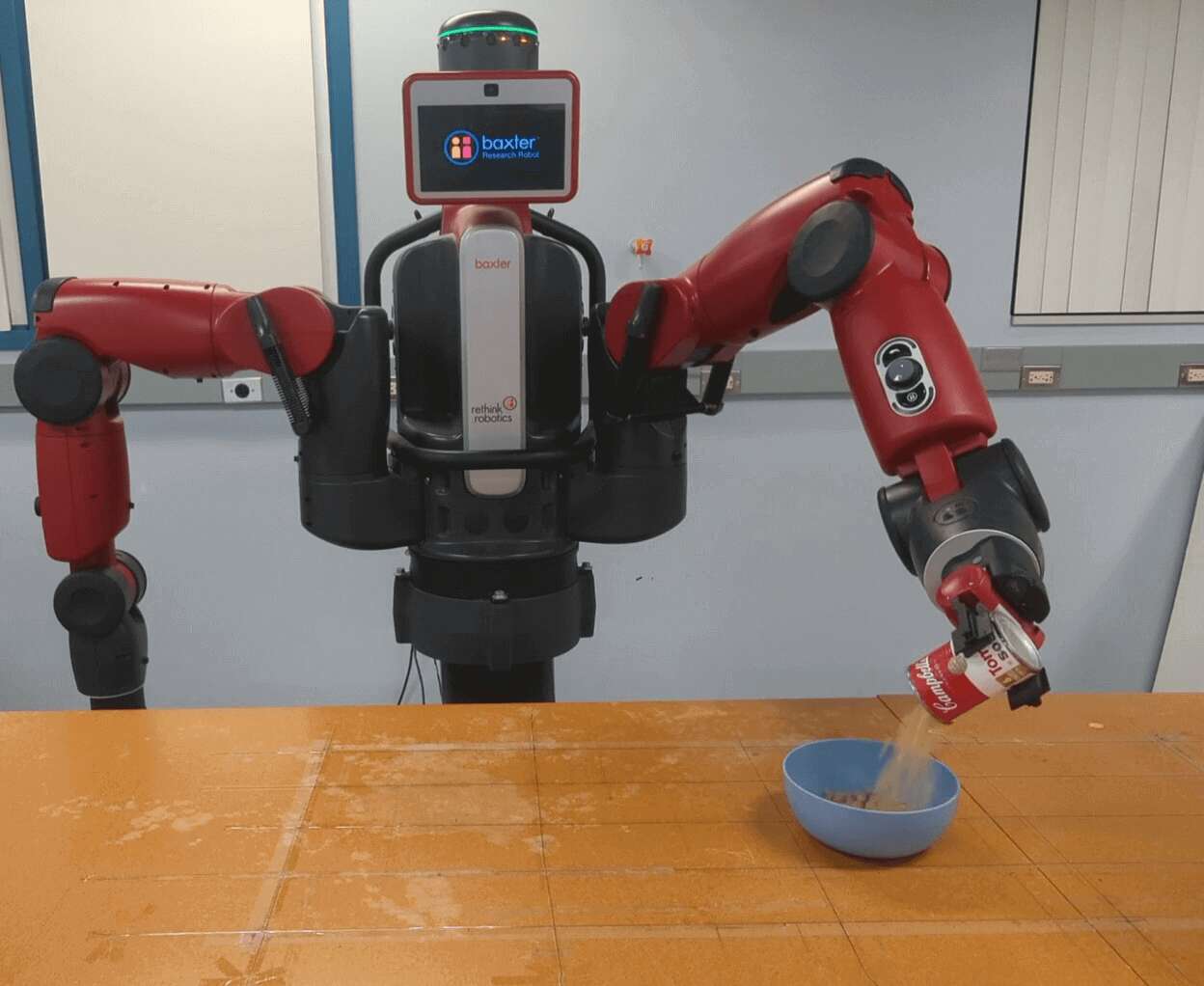} & \includegraphics[width=0.15\textwidth]{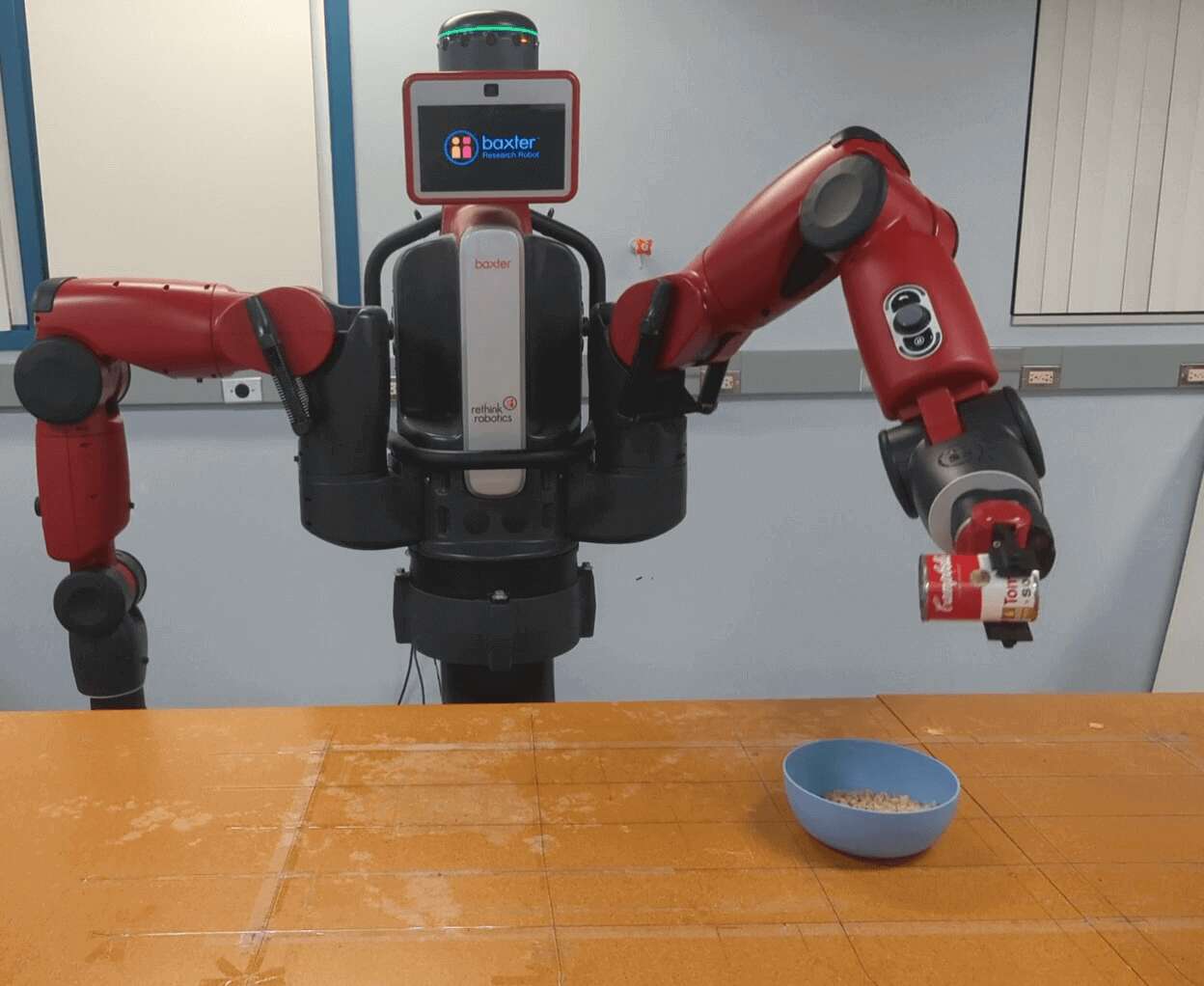}\\
            \multicolumn{5}{c}{\footnotesize{Execution from $\id{soup\_can}$ to $\id{bowl}$}}\\
            \includegraphics[width=0.15\textwidth]{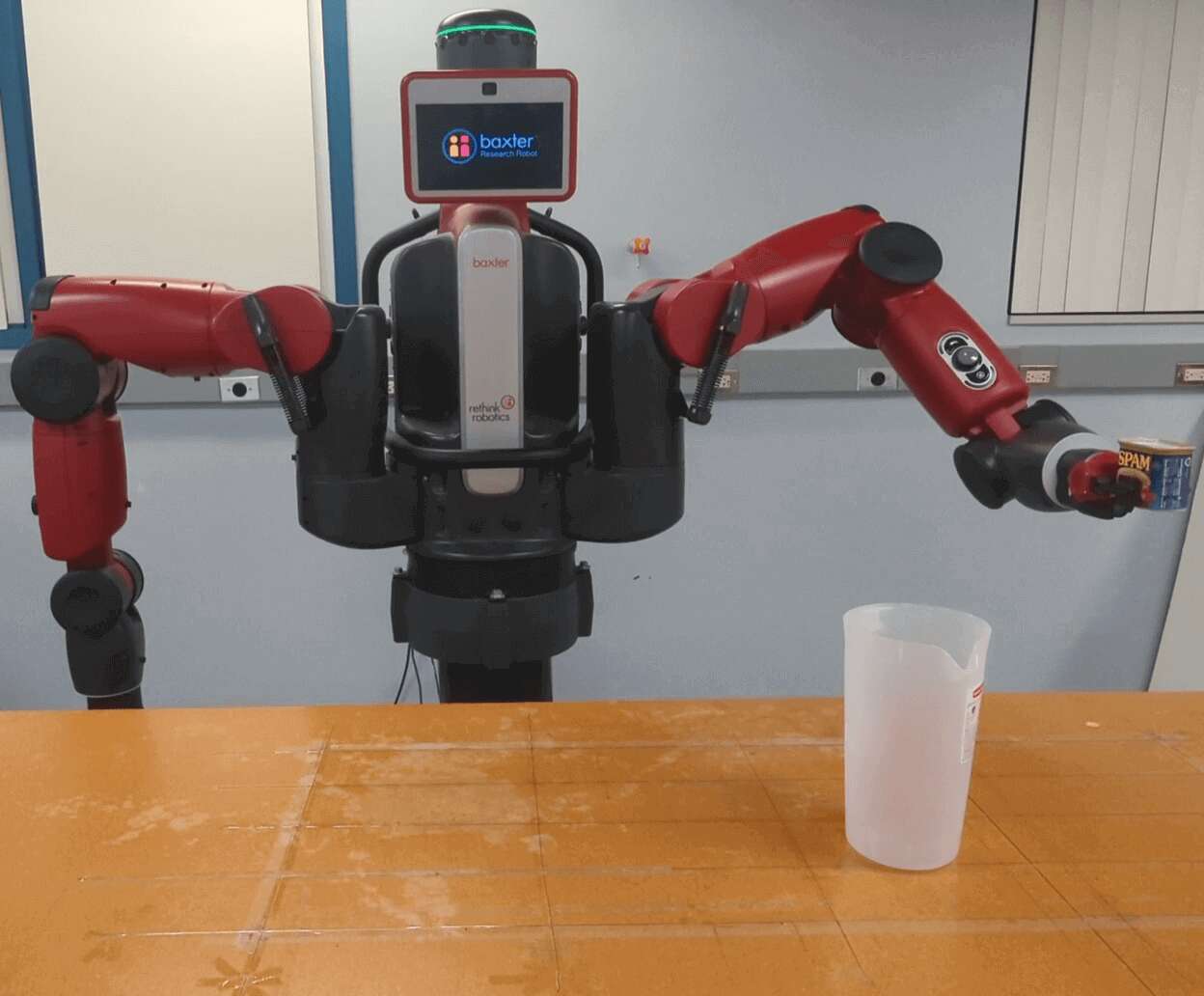} & \includegraphics[width=0.15\textwidth]{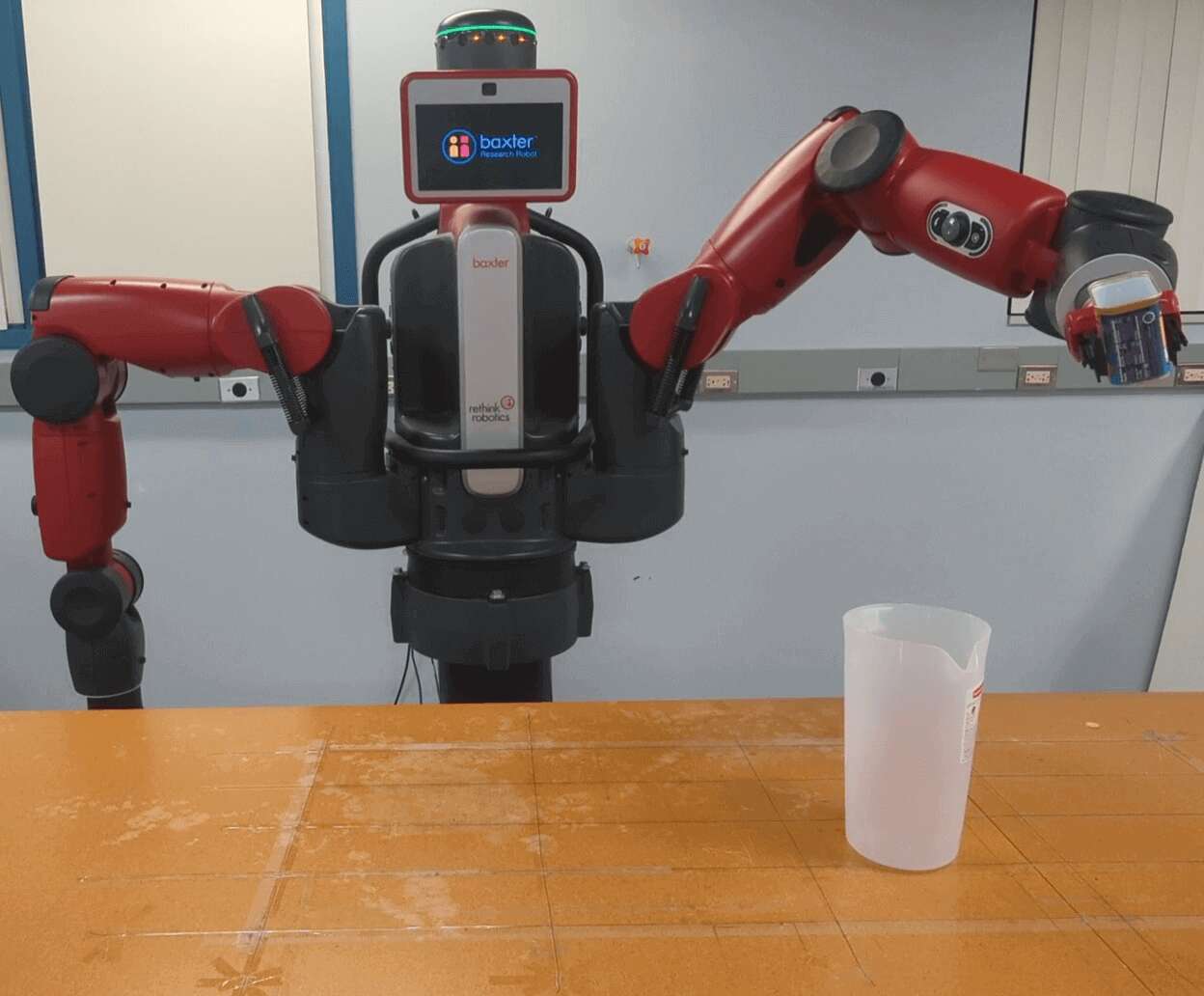} & \includegraphics[width=0.15\textwidth]{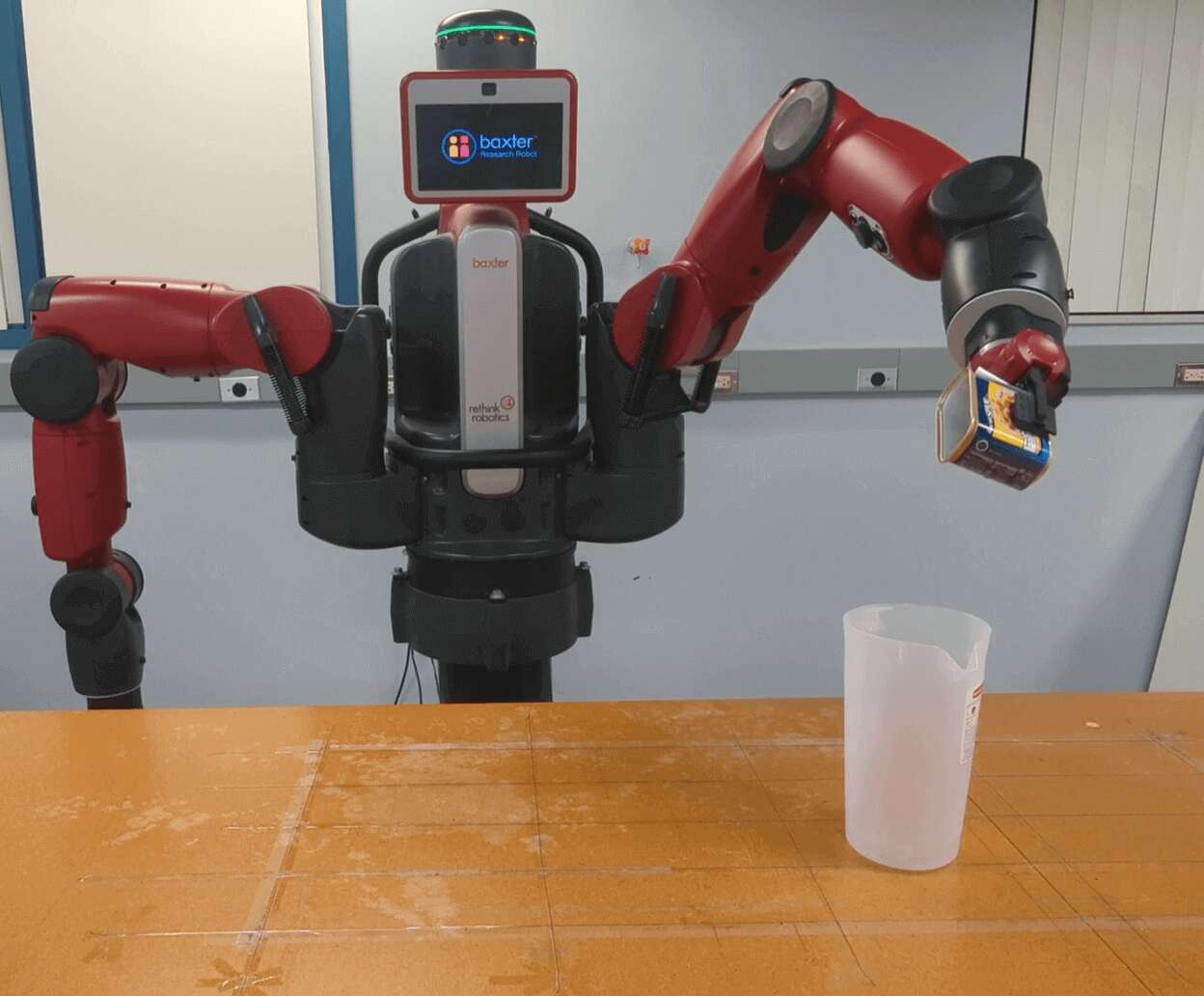} & \includegraphics[width=0.15\textwidth]{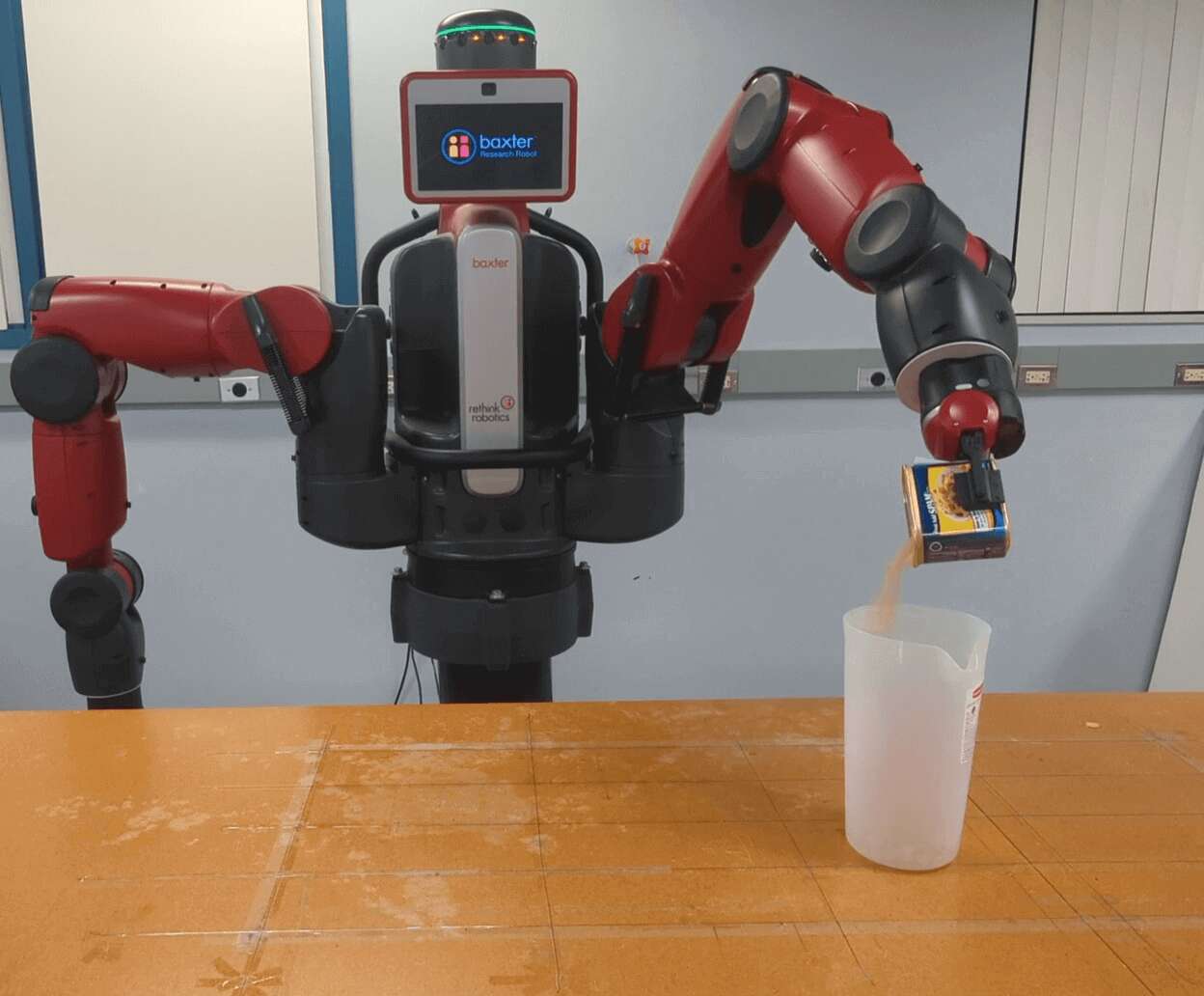} & \includegraphics[width=0.15\textwidth]{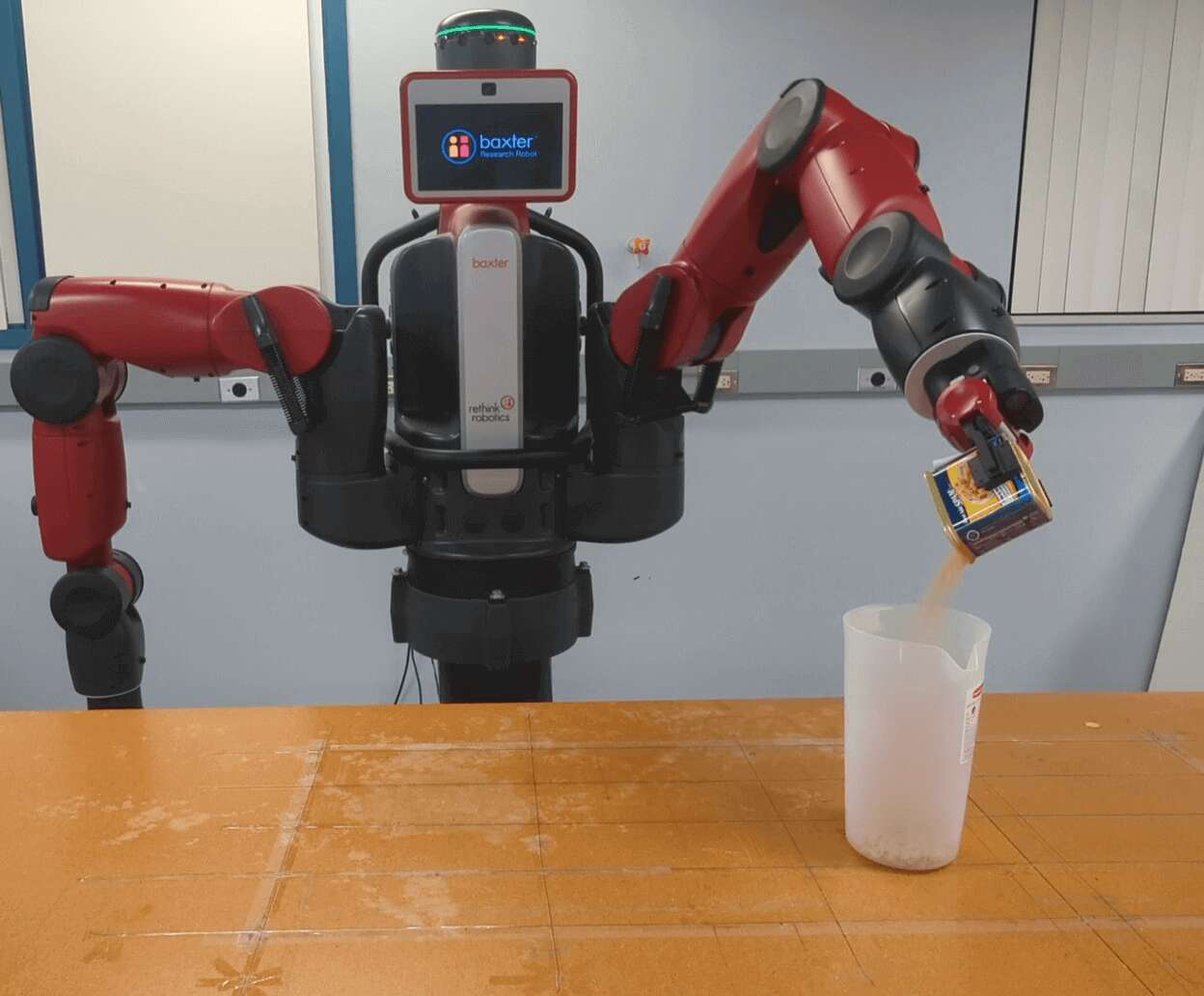}\\
            \multicolumn{5}{c}{\footnotesize{Execution from $\id{Spam\_can}$ to $\id{pitcher}$}}\\
            \includegraphics[width=0.15\textwidth]{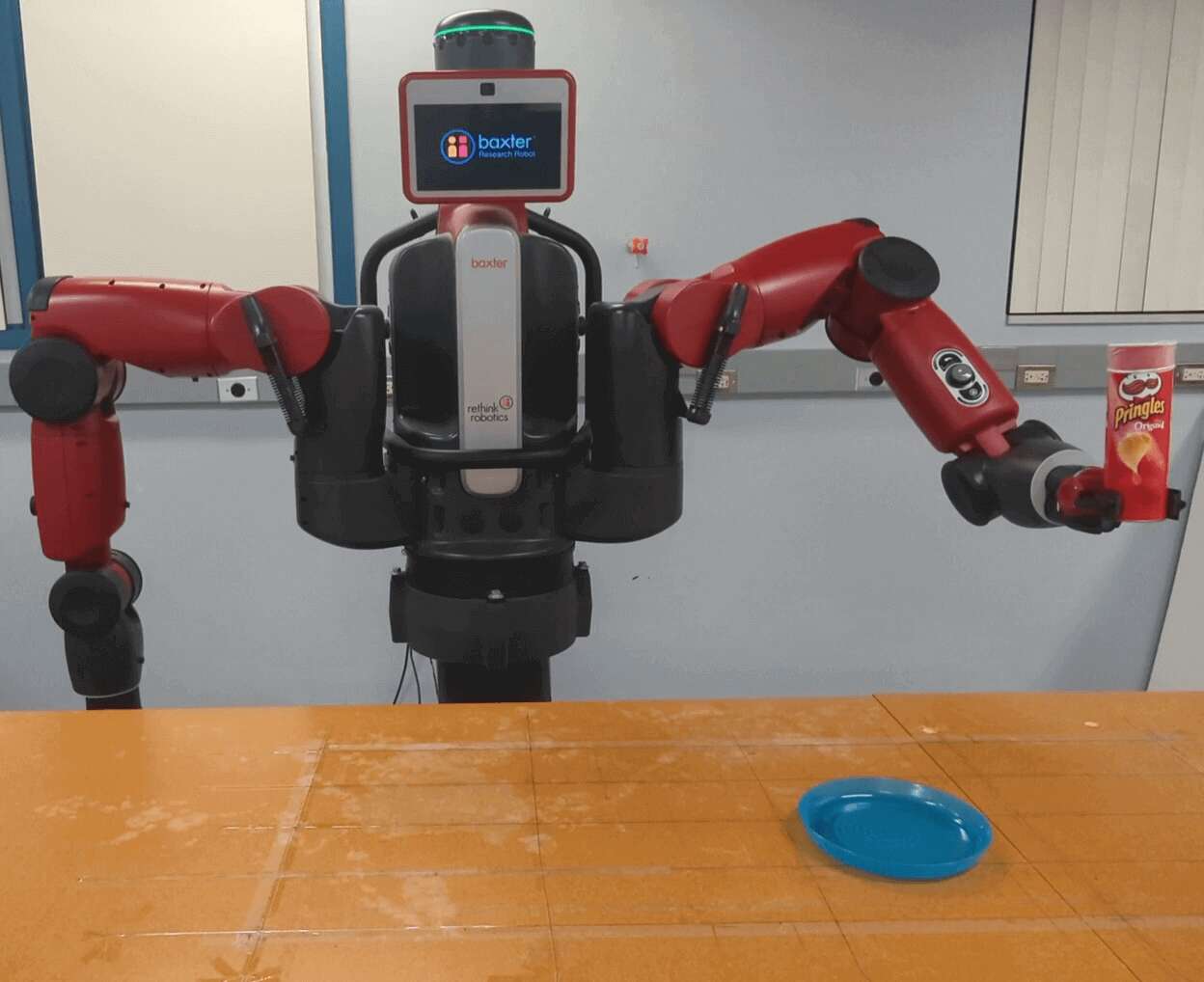} & \includegraphics[width=0.15\textwidth]{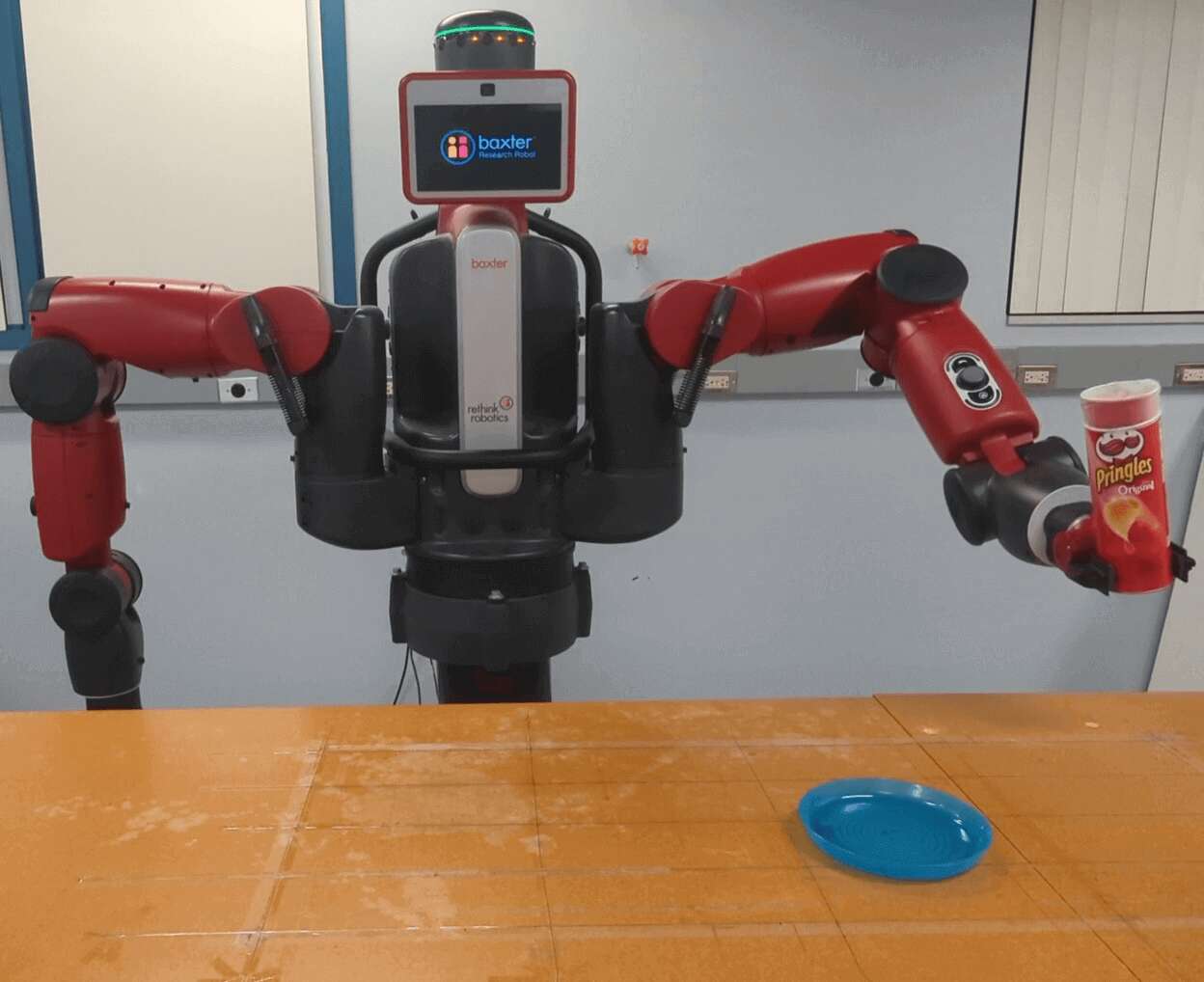} & \includegraphics[width=0.15\textwidth]{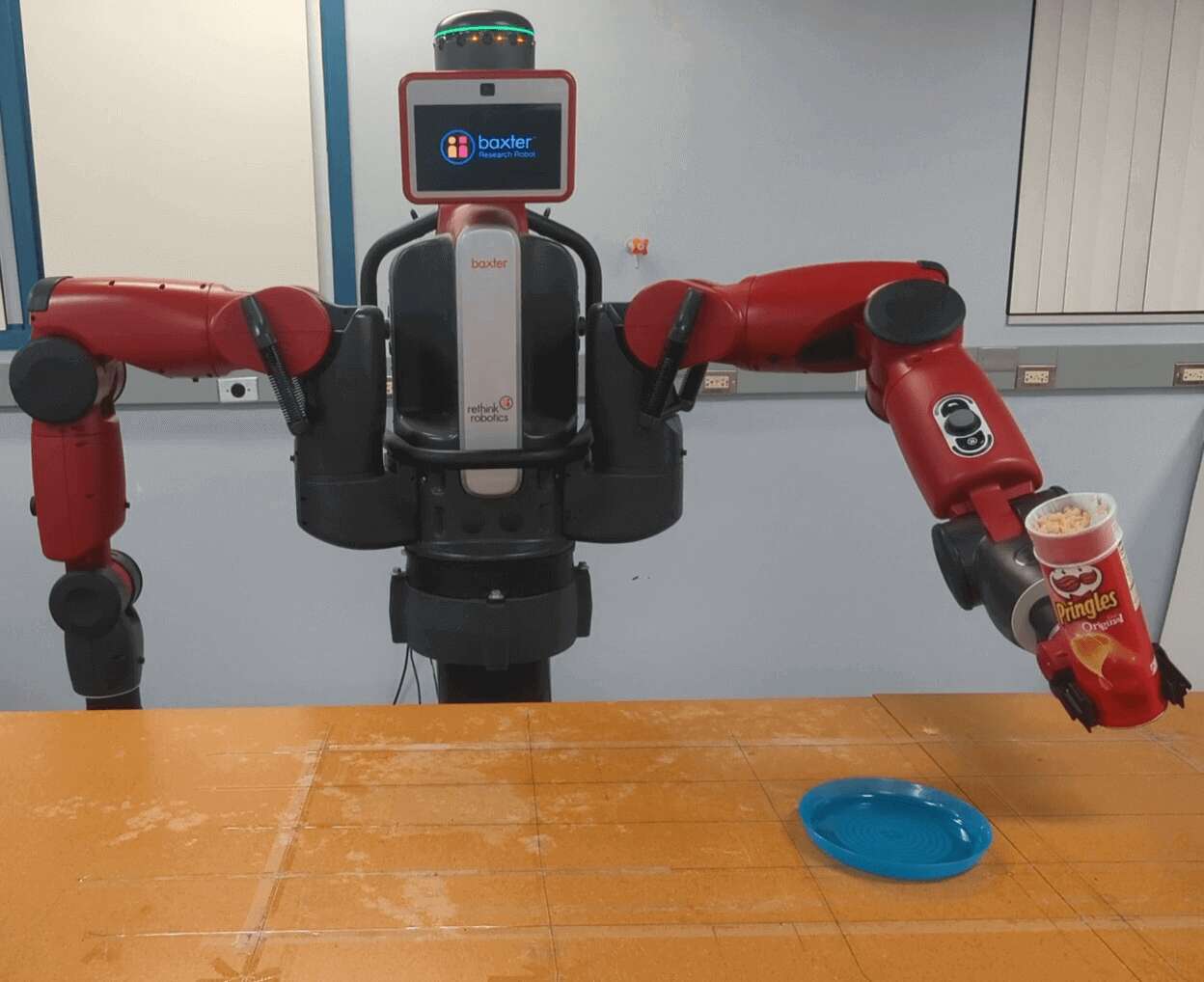} & \includegraphics[width=0.15\textwidth]{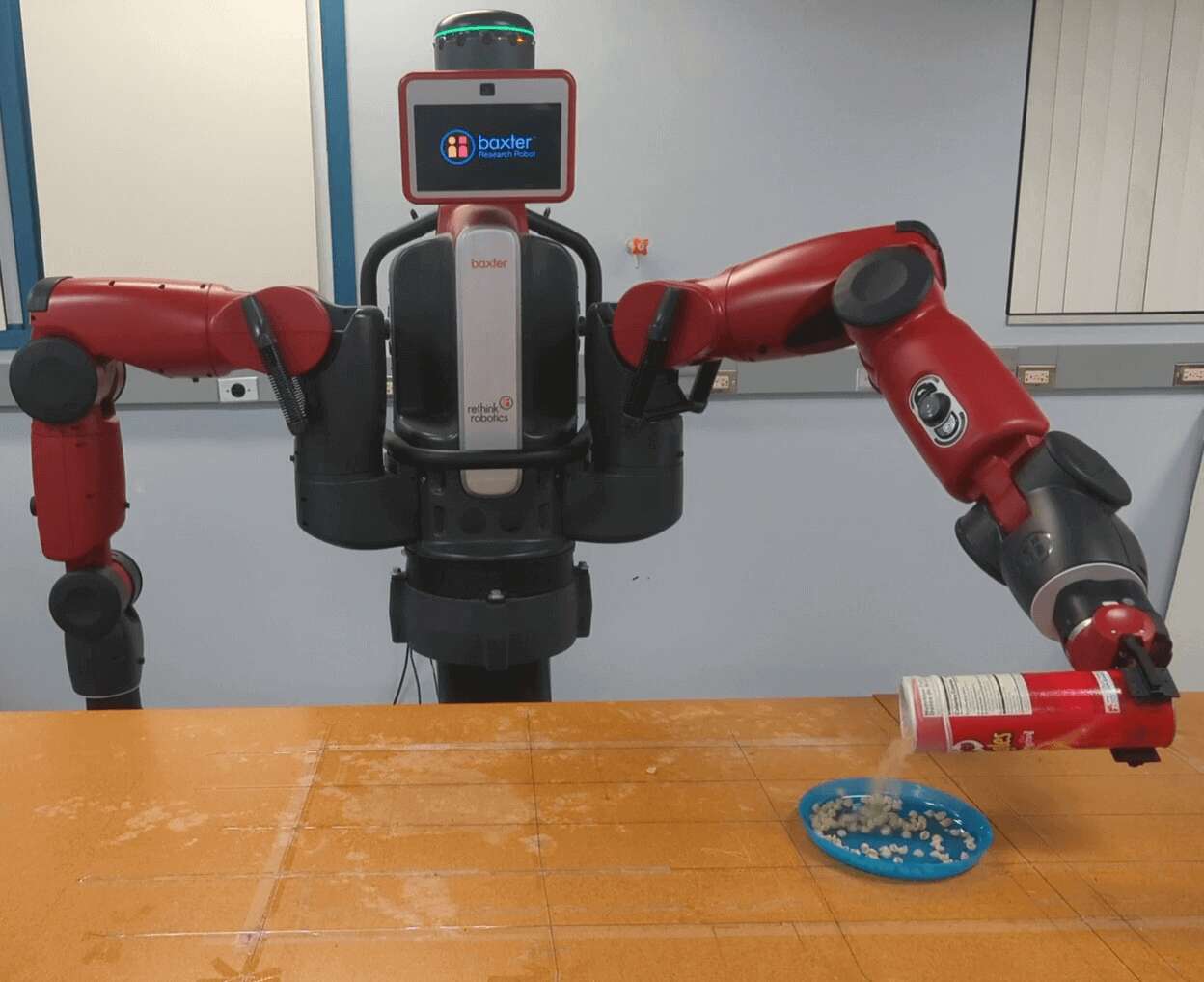} & \includegraphics[width=0.15\textwidth]{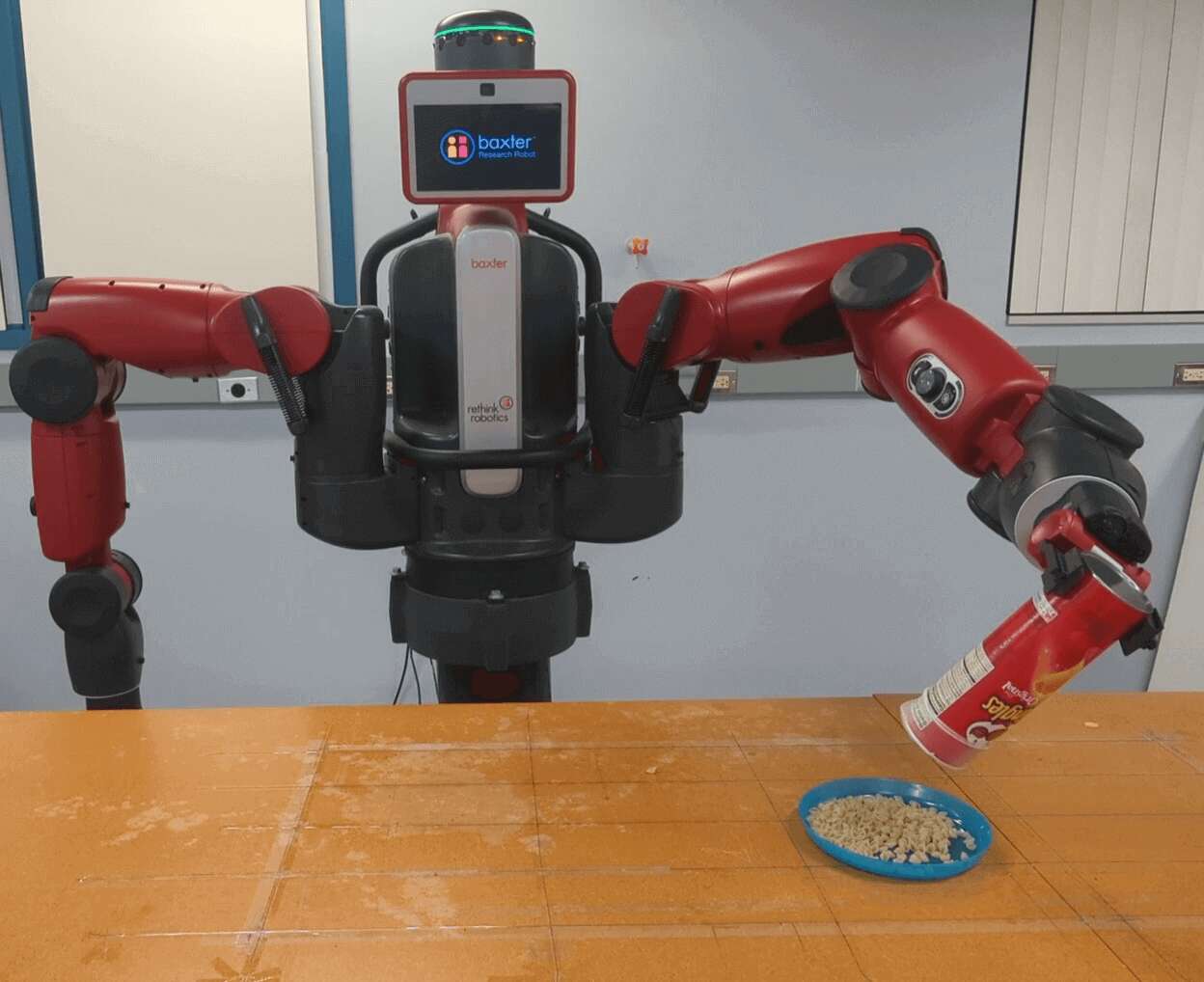}\\
            \multicolumn{5}{c}{\footnotesize{Execution from $\id{Pringles\_box}$ to flat $\id{plate}$}}
        \end{tabular}
        \captionsetup{labelformat=empty}
        \caption{Three task executions using \textcolor{red}{demonstration \#2}}
    \end{subfigure}
    \par\bigskip
    \caption{Sample task executions -- $3$ for each demonstration. For each execution, $5$ sequential frames are shown from left to right.}
    \label{fig:hardware_trials}
    \vspace{-0.25\baselineskip}
\end{figure*}

\subsubsection{Motion transfer using the transformed guiding poses and the \mbox{$\left\{\relframe{C}{}{}{}{}\right\}$} frames of the new task-relevant objects}

The final step is to transfer each guiding pose in $\relframe{{\bm{\Gamma}}}{}{d}{}{C_sC_r}$ to capture the pose of \mbox{$\left\{\relframe{C}{}{n}{}{r}\right\}$} represented w.r.t. the pose of \mbox{$\left\{\relframe{C}{}{n}{}{s}\right\}$}. This ultimately enables us to compute a new sequence of \emph{guiding poses} $\relframe{{\bm{\Gamma}}}{}{n}{}{}$, i.e. end-effector poses, from the sequence of original \emph{guiding poses} $\relframe{{\bm{\Gamma}}}{}{d}{}{}$, extracted from the demonstration (see \S\ref{sec:demonstration_representation}).

In order to transfer the relative poses of the \emph{motion-transfer} frames extracted from the demonstration to the new task instance, each relative pose $\relframe{\mathbf{g}}{}{d}{(i)}{C_sC_r} \in \relframe{{\bm{\Gamma}}}{}{d}{}{C_sC_r} \forall i\in [1,k]$, between the \emph{motion-transfer} frames of the primary \mbox{$\left(\left\{\relframe{C}{}{d}{}{r}\right\}\right)$} and the passive \mbox{$\left(\left\{\relframe{C}{}{d}{}{s}\right\}\right)$} objects used in the demonstration, is transferred to generate the relative transformation of the frame \mbox{$\left\{\relframe{C}{}{n}{}{r}\right\}$}, of the primary object represented in the frame \mbox{$\left\{\relframe{C}{}{n}{}{s}\right\}$}, of the passive object in the new task instance $t_n$. This new sequence is denoted by
\begin{equation}
\begin{aligned}
\label{eq:task_instance_c_frame}
    \relframe{{\bm{\Gamma}}}{}{n}{}{C_r} = \left\langle \relframe{\mathbf{g}}{}{n}{(1)}{C_r}, \cdots, \relframe{\mathbf{g}}{}{n}{(k)}{C_r} \right\rangle
\end{aligned}
\end{equation}
where, $\relframe{\mathbf{g}}{}{n}{(i)}{C_r} = \relframe{\mathbf{g}}{}{n}{}{C_s} \relframe{\mathbf{g}}{}{d}{(i)}{C_sC_r} \quad \forall i\in [1,k]$

The process completes with the construction of $\relframe{{\bm{\Gamma}}}{}{n}{}{}$, the sequence of new \emph{guiding poses} i.e. end-effector poses for the new task instance, expressed in the \emph{motion-transfer} frame \mbox{$\left\{\relframe{C}{}{n}{}{s}\right\}$} of its passive object. $\relframe{{\bm{\Gamma}}}{}{n}{}{}$ is computed from the sequence $\relframe{{\bm{\Gamma}}}{}{n}{}{C_r}$ as follows.
\begin{equation}
\begin{aligned}
\label{eq:end_effector_c_frame}
    \relframe{{\bm{\Gamma}}}{}{n}{}{} = \left\langle \relframe{\mathbf{g}}{}{n}{(1)}{E}, \cdots, \relframe{\mathbf{g}}{}{n}{(k)}{E} \right\rangle
\end{aligned}
\end{equation}
where, $\relframe{\mathbf{g}}{}{n}{(i)}{E} = \relframe{\mathbf{g}}{}{n}{(i)}{C_r} \left(\relframe{\mathbf{g}}{}{n}{}{EB_r} \relframe{\mathbf{g}}{}{n}{}{B_rC_r}\right)^{-1} \qquad\forall i\in [1,k]$

Using the new guiding poses in $\relframe{{\bm{\Gamma}}}{}{n}{}{}$, the ScLERP-based motion planner with Jacobian pseudo-inverse~\cite{sarker2020screw} is used to compute the motion plan in the joint space from an initial configuration (see \S\ref{sec:demonstration_representation}).

\mypara{Computational Criterion for Task Completion}
To ensure that the robot has successfully completed the task, we also need a computational criterion for task completion. For pouring, let $(a_s, b_s, n_s)$ be the parameters of the superellipse that represents the rim of the passive object. At the configuration of the arm $\relframe{{\bm{\theta}}}{}{n}{(i)}{}$, let $\left(p_x^{(i)}, p_y^{(i)}\right)$ be the projection of the origin of \mbox{$\left\{\relframe{C}{}{n}{}{r}\right\}$} on the plane containing the rim of the passive object, expressed in a reference frame in which the rim of the passive object is expressed as $\left|\frac{x}{a_s}\right|^{n_s} + \left|\frac{y}{b_s}\right|^{n_s} = 1$. Then for points in the planned path for the new task instance with index $i \geq i_0$, if $\left|\frac{p_x^{(i)}}{a_s}\right|^{n_s} + \left|\frac{p_y^{(i)}}{b_s}\right|^{n_s} < 1$, the contents will be poured in. The index $i_0$ can be calculated based on \emph{angle of tilt} of the primary object and would depend on \emph{height of the content} within the primary object.


\section{Experimental Results}


\subsubsection{Results of Kinematic Simulation}

The values of $(a,b,n,h)$ for the primary and passive objects used in the demonstration (\textcolor{red}{red} in Fig.~\ref{fig:simulation}) were $(3.25,3.25,2,10)$ and $(8,8,2,5.5)$, respectively, with all lengths in cm. Using these values, we grouped the virtual primary and passive objects of the new task instances (\textcolor{blue}{blue} in Fig.~\ref{fig:simulation}) into $4$ categories: \emph{Fat Tall}, \emph{Fat Short}, \emph{Thin Tall}, and \emph{Thin Short} -- resulting in a total of $16$ groups.

From each group, we selected $200$ virtual primary-passive object pairs (i.e. $200$ task instances), $20$ for each value of $n \in \{0.5, 0.7, 1, 1.5, 2, 2.5, 4, 5, 7, 8\}$, uniformly sampled at different locations on the table. Using the physical demonstration, we generated two successful motion plans (that did not violate joint limits) for each task instance -- one with the \emph{motion-transfer} frames \mbox{$\left\{\relframe{C}{}{}{}{}\right\}$} and the other using the baseline planner~\cite{mahalingam2023human} (w/o the \emph{motion-transfer} frames). Table~\ref{tab:simulation_result} shows that our approach significantly outperforms the baseline approach. Numbers in \textcolor{blue}{blue} and \textcolor{red}{red} respectively denote the percentage of collision-free executions with and without the \emph{motion-transfer} frames \mbox{$\left\{\relframe{C}{}{}{}{}\right\}$}. 

Since task constraints are not explicit, it is challenging to accurately characterize the successful completion of the task, in general. For the specific task of pouring, one measure of the execution quality of the task is \emph{maximum vertical tilt angle} along the path, when the motion transfer frame of the primary object lies outside the rim of the passive object. Note that smaller angles indicate that a larger volume of liquid can be carried and poured without spilling the contents outside the passive object (the intended pouring target). 

Each cell in Table~\ref{tab:simulation_result1} shows the maximum and minimum of the maximum vertical tilt angles across $150$ randomly sampled task instances (with collision-free paths) for our technique (\textcolor{blue}{blue}) and the baseline (\textcolor{red}{red})~\cite{mahalingam2023human}. The results show that the \emph{range} of the maximum vertical tilt angles using our technique is narrower than that of the baseline. Furthermore, the maximum value of the maximum vertical tilt angles for our technique is substantially smaller than or comparable to the baseline, indicating that our approach is more robust compared to the baseline. However, a head-to-head comparison of the maximum vertical tilt angle for individual task instances shows that in many cases ($1239$ out of $2400$) the maximum vertical tilt angle of our technique is higher than that of the baseline. The difference was usually small and never more than $11^\circ$ from that of the baseline. Nevertheless, as discussed above, the baseline could be substantially worse than our technique and in many task instances by a large margin of $90^\circ$ or higher.

Fig.~\ref{fig:simulation} shows some of the successful executions of tasks in the kinematic simulation with objects of varying shapes and sizes. The demonstration (with the red $\id{soup\_can}$) was acquired physically and transferred to the simulation; whereas for the new task instances, virtual objects (the blue and white containers) of varying geometries were generated by changing the parameters $(a,b,n,h)$. 

\begin{table*}[!h]
    \centering
    \begin{threeparttable}
        \begin{tabular}{c|c|c|c|c}
            \backslashbox[2.6cm]{Passive}{Primary} & \emph{Fat Tall} & \emph{Fat Short} & \emph{Thin Tall} & \emph{Thin Short} \\
            \midrule
            \emph{Fat Tall} & 
            [\textcolor{blue}{15.9, 27.2}] vs. [\textcolor{red}{12.4, 26.5}] & [\textcolor{blue}{17.0, 27.2}] vs. [\textcolor{red}{13.9, 38.4}] & [\textcolor{blue}{17.3, 27.2}] vs. [\textcolor{red}{15.1, 31.7}] & [\textcolor{blue}{17.2, 27.2}] vs. [\textcolor{red}{16.0, 88.4}] \\
            \emph{Fat Short} & 
            [\textcolor{blue}{16.2, 27.2}] vs. [\textcolor{red}{12.5, 28.1}] & [\textcolor{blue}{14.0, 28.4}] vs. [\textcolor{red}{11.6, 48.1}] & [\textcolor{blue}{16.4, 28.4}] vs. [\textcolor{red}{14.3, 31.7}] & [\textcolor{blue}{17.2, 28.4}] vs. [\textcolor{red}{16.1, 127.1}] \\
            \emph{Thin Tall} & 
            [\textcolor{blue}{24.1, 39.5}] vs. [\textcolor{red}{20.4, 127.1}] & [\textcolor{blue}{24.1, 39.5}] vs. [\textcolor{red}{22.7, 127.1}] & [\textcolor{blue}{24.1, 39.5}] vs. [\textcolor{red}{22.9, 52.9}] & [\textcolor{blue}{24.1, 41.6}] vs. [\textcolor{red}{32.9, 127.1}] \\
            \emph{Thin Short} & 
            [\textcolor{blue}{24.1, 39.5}] vs. [\textcolor{red}{20.4, 127.1}] & [\textcolor{blue}{24.1, 39.5}] vs. [\textcolor{red}{22.9, 127.1}] & [\textcolor{blue}{24.1, 39.5}] vs. [\textcolor{red}{22.7, 54.2}] & [\textcolor{blue}{24.1, 39.5}] vs. [\textcolor{red}{31.7, 127.1}] \\
            \bottomrule
        \end{tabular}
        \caption{Range of \emph{maximum vertical tilt angles} (in degrees) of the primary object while outside the passive one -- in our approach (\textcolor{blue}{blue}) vs. the baseline planner (\textcolor{red}{red})~\cite{mahalingam2023human}.}
        \label{tab:simulation_result1}
    \end{threeparttable}
\end{table*}

\subsubsection{Results of Hardware Experiments}
All physical experiments were conducted using the Baxter robot from Rethink Robotics~\cite{baxter_hardware_specs}. For experiments, we considered the following objects: $\id{soup\_can},$ $\id{bowl},$ $\id{plate},$ $\id{Spam\_can},$ $\id{coffee\_can},$ $\id{pan},$ $\id{CheezIt\_box},$ $\id{DominoSugar\_box},$ $\id{pitcher},$ $\id{Pringles\_can}$ from the YCB benchmark~\cite{calli2017yale}. The $\id{bowl},$ $\id{pan},$ $\id{plate},$ and the $\id{pitcher}$ were used as $\emph{passive}$ objects, and the rest as $\emph{primary}$ objects. We provide two kinesthetic demonstrations (see Fig.~\ref{fig:hardware_demonstration}) for the \emph{pouring} task --
\begin{enumerate*}[label=demonstration \#\theenumi:,leftmargin=10\parindent]
    \item from $\id{soup\_can}$ to red $\id{bowl}$, and
    \item from $\id{CheezIt\_box}$ to flat blue $\id{plate}$.
\end{enumerate*}

We performed a total of $40$ trials with $2$ demonstrations ($20$ trials for each) varying the objects and their poses.
Fig.~\ref{fig:hardware_trials} shows $6$ sample task executions on different sets of objects using the two demonstrations.
We did not observe execution failures due to collisions between objects or due to joint limit violations. However, in a few (3-4) executions, a small fraction of the granular material (pasta and rice) spilled outside. These results reinforce our overall claim of generating a robust manipulation plan for a task without requiring a direct demonstration on its participating objects.
\section{Conclusion}
In this paper, we presented a framework for computing motion plans for complex manipulation tasks across a variety of functionally similar but geometrically different objects based on kinesthetic demonstrations. We showed that by appropriate choice of critical points relevant to the task and assignment of reference frames to these points, we can extend a screw-geometric framework for manipulation planning from demonstration~\cite{mahalingam2023human} to different objects. Using the example task of pouring, we presented both simulation and experimental results, showing that our approach can indeed allow for planning across a variety of objects. Future work will involve further experimentation on complex manipulation tasks such as scooping and stacking.


\bibliographystyle{IEEEtran}
\bibliography{c_frame}

\end{document}